\documentclass[lettersize,journal]{IEEEtran}
\usepackage{hyperref}
\usepackage{algorithmic}
\usepackage{algorithm}
\usepackage{array}
\usepackage{textcomp}
\usepackage{stfloats}
\usepackage{url}
\usepackage{verbatim}
\usepackage{graphicx}
\usepackage{cite}
\usepackage{multirow} 
\usepackage{multirow}
\usepackage{makecell}

\usepackage{tikz}
\usepackage{amsmath}
\usepackage{enumitem}
\usepackage{makecell}
\usepackage{booktabs}
\usepackage{graphicx}
\usepackage{caption}
\usepackage{subcaption}
\usepackage{xpatch}
\usepackage{xcolor}
\newcommand{\rev}[1]{#1}
\newcommand{\revision}[1]{\textcolor{black}{#1}}
\newcommand{\new}[1]{\textcolor{black}{#1}}
\newcommand{\delete}[1]{\textcolor{black}{#1}}
\newcommand{\del}[1]{\textcolor{gray}{#1}}

\AtBeginDocument{%
  }
\newcommand{\eg}{\textit{e.g.}}
\newcommand{\ie}{\textit{i.e.}}
\newtheorem{observation}{\textbf{Implication}}
\DeclareRobustCommand*{\IEEEauthorrefmark}[1]{%
  \raisebox{0pt}[0pt][0pt]{\textsuperscript{\footnotesize #1}}%
}
\begin{document}

\title{Understanding Large Language Models in Your Pockets: Performance Study on COTS Mobile Devices}

\author{\author{
	\IEEEauthorblockN{
		Jie
        Xiao\IEEEauthorrefmark{1}, 
		Qianyi
        Huang\IEEEauthorrefmark{1}, 
		Xu
        Chen\IEEEauthorrefmark{1}, 
		Chen 
        Tian\IEEEauthorrefmark{2}} 
	\\
    \IEEEauthorblockA{\IEEEauthorrefmark{1}Sun Yat-Sen University}\\
	\IEEEauthorblockA{\IEEEauthorrefmark{2}Nanjing University}
}

\thanks{This paper was produced by the IEEE Publication Technology Group. They are in Piscataway, NJ.}%
\thanks{Manuscript received April 19, 2021; revised August 16, 2021.}}

\markboth{Journal of \LaTeX\ Class Files,~Vol.~14, No.~8, August~2021}%
{Shell \MakeLowercase{\textit{et al.}}: Understanding Large Language Models in Your Pockets: Performance Study on COTS Mobile Devices}

\maketitle

\begin{abstract}
As large language models (LLMs) increasingly integrate into every aspect of our work and daily lives, there are growing concerns about user privacy, which push the trend toward local deployment of these models. There are a number of lightweight LLMs (e.g., Gemini Nano, LLAMA2 7B) that can run locally on smartphones, providing users with greater control over their personal data. As a rapidly emerging application, we are concerned about their performance on commercial-off-the-shelf mobile devices. To fully understand the current landscape of LLM deployment on mobile platforms, we conduct a comprehensive measurement study on mobile devices. %

\rev{While user experience is the primary concern for end-users, developers focus more on the underlying implementations. Therefore, we evaluate \new{both user-centric metrics—such as token throughput, latency, and response quality—and developer-critical factors, including resource utilization, OS strategies, battery consumption, and launch time.}}
We also provide comprehensive comparisons across the mobile system-on-chips (SoCs) from major vendors, highlighting their performance differences in handling LLM workloads, \rev{which may help developers identify and address bottlenecks for mobile LLM applications}. We hope that this study can provide insights for both the development of on-device LLMs and the design for future mobile system architecture.
\end{abstract}


\section{Introduction}
\IEEEPARstart{O}{ver} the past two years, Large Language Models (LLMs) have demonstrated exceptional performance across a wide range of domains. Their advanced capabilities in understanding human language and generating text have made them indispensable in various applications, including dialogue systems, machine translation, and text generation. Several companies have begun integrating LLMs as system services (e.g., Apple Intelligence~\cite{appleintelligence}, 
Copilot+ PC~\cite{copilot}, where these models assist users in organizing daily routines, summarizing emails and messages, and even generating automated replies. However, while LLMs serve as powerful and appealing personal assistants, these models continuously monitor and analyze users' activities, which leads to serious privacy concerns~\cite{he2024emerged}. 

To address the privacy concerns, local deployment of LLMs is becoming attractive, as it allows these models to perform inference directly on the device and no personal data are sent to cloud servers. In this way, users have greater control of their personal data. While LLMs, such as ChatGPT~\cite{chatgpt} and Claude~\cite{claude},
are typically deployed on the cloud, local deployment is becoming feasible for two reasons. On one hand, \new{with the development of resource-efficient strategies~\cite{10.1145/3706418}}, a series of lightweight LLMs (e.g., Llama2-7B~\cite{touvron2023llama} and Mistral-7B~\cite{jiang2023mistral}) have been introduced. These models, with billions of parameters, are significantly smaller in size compared to cloud-based models, which typically have hundreds of billions of parameters. Despite their reduced size, these lightweight LLMs maintain a high level of expressiveness in both common-sense reasoning and domain-specific knowledge, and thus can fully satisfy the basic needs of mobile users. On the other hand, state-of-the-art mobile SoCs demonstrate significant improvement in computing capability and memory capacity. More importantly, a variety of specialized AI accelerators are integrated into the mobile processors, such as Graphics Processing Units (GPUs) and Neural Processing Units (NPUs). These hardware improvements make it possible to implement complex AI functions on mobile devices.

As a rapidly emerging application, LLMs differ significantly from traditional mobile applications in terms of computational complexity and memory demands~\cite{yin2024llmservicemobiledevices}. Therefore, we are particularly concerned with their performance on commercial off-the-shelf (COTS) mobile devices, as well as the performance fluctuations caused by mobile operating systems and hardware architecture compatibility.
Several studies have taken the first steps to explore the current state of mobile LLM\cite{laskaridis2024melting}\cite{li2024large}\cite{murthy2024mobileaibench}. They have conducted preliminary assessments of model deployment and inference performance through the development of specialized benchmark suites.
However, they mainly focus on performance metrics such as token throughput and latency, without looking into the underlying hardware dynamics and system configurations, such as CPU/GPU utilization, operating frequencies, etc. 
Besides, there is a lack of analysis about how these hardware and system-level factors affect on-device LLM inference. This is essential for developers to identify and address bottlenecks for mobile LLM applications.

In this paper, we aim to reveal the current state and potential of mobile hardware for supporting on-device LLMs by deploying LLMs across a diverse range of mobile devices. We select multiple high-tier SoCs from different vendors that represent currently available resources and development trends in mobile hardware. We analyze the differences in SoCs from major vendors and ultimately select SoCs represented by Qualcomm, HiSilicon, MediaTek and \revision{Apple} to perform local LLM inference. \new{We deploy six models from various families (Llama$2$/$3.2$~\cite{llama3.2}, Qwen$3$~\cite{yang2025qwen3}, Gemma$3$~\cite{team2025gemma})}
on mobile devices with llama.cpp~\cite{llamacpp} and MLC LLM~\cite{mlc-llm}, which are the two popular mobile LLM inference engines. During inference, we collect comprehensive metrics with specific profilers including Snapdragon Profiler~\cite{snapdragon}, Arm Streamline~\cite{streamline} \revision{and Xcode~\cite{Xcode}} to make sure that all the data accurately reflect the hardware performance. We investigate how the hardware specifications, including big.LITTLE cores\cite{armbig.little} and \new{accelerators' architecture}, affect the inference performance, which is essential to help the development of future mobile SoCs to accelerate LLMs.

Considering the diversity of mobile hardware and rapid iteration of capabilities, current research exhibits a notable gap in both comprehensive understanding and detailed evaluation of LLM deployment on mobile devices. In order to show the performance of on-device inference and provide developers with valuable insights, we have summarized the following key questions:
        
        \noindent\textbf{Q1. Hardware Requirements}: What are the key hardware requirements (e.g., computation capability, RAM size) to support LLMs on mobile devices?

        \noindent\new{\textbf{Q2. Model and Quantization}: What is the trade-off between model accuracy and throughput across different quantization schemes? How does model architecture influence sensitivity to quantization?}
        
        \noindent\textbf{Q3. Resource Configuration}: What is the optimal runtime resource configuration, including system policies (DVFS \revision{and thermal throttling}), the number of parallel threads, core affinity settings, and others? \revision{Moreover, how does the optimal configuration differ when inference runs exclusively versus concurrently with other tasks?} Furthermore, how can this configuration be adapted to different devices? 
        
        \noindent\textbf{Q4. Software-Hardware Alignment}: Do the implementations and optimizations of inference frameworks align with general-purpose processors (CPUs) or specialized accelerators (GPUs) from various vendors? How much do software implementations limit the utilization of hardware capabilities, such as computation throughput or memory bandwidth?

From our study, we have the summarized corresponding observations to the above questions:

\textbf{Response to Q1:}
    \begin{itemize}
        \item \textit{CPU performance}: \revision{The iOS device shows a clear advantage, particularly in the prefill stage. In the decode stage, while the Apple M1 reaches about 10 tokens/s, non-iOS devices with Armv8-A CPUs (e.g., Cortex-A76 and A77) only achieve 2–4 tokens/s. In contrast, Armv9-A CPUs (e.g., Cortex-X4), equipped with specialized instructions such as smmla and sdot, deliver substantial gains and can reach throughput comparable to iOS devices.}
        \item 
        \textit{\new{Performance of hardware accelerators}}: \new{As for accelerators, 
        GPU demonstrates promising preliminary acceleration in the prefill stage, while this speedup is limited in the memory-bound decode stage. At this stage, GPU acceleration support is available only for a limited set of devices.}
        \item 
        \textit{RAM requirement}: The memory (RAM) requirement to support LLMs depends on the model size and the quantization bit-width. \new{For most mobile devices, a $4$-bit quantized $7$B-parameter model is the upper limit. It typically requires at least $4$ GB of RAM to store the model weights and maintain a minimal context (\eg, KV cache).}
        \item \textit{Battery support}: \new{For CPU-based inference with a $64$-token prompt and $128$-token generation on modern high-tier mobile devices, %
        power consumption is approximately $20$ mAh per round. Given that typical smartphone batteries have a capacity of $4000$ mAh to $6000$ mAh, these devices can support hundreds of rounds of inference. Instruction acceleration for CPU inference further improves power efficiency. Notably, GPU-based inference provides a significant energy efficiency advantage over CPU execution, highlighting its potential for optimized performance on mobile devices.}
    \end{itemize}

\textbf{Response to Q2:}
    \begin{itemize}
    \item \new{\textit{Trade-off between accuracy and throughput}: 4-bit quantization strikes the optimal balance by achieving peak or near-peak speeds in both prefill and decoding stages while maintaining robust inference accuracy. For small-scale LLMs, such as Gemma$3$-$1$B, $8$-bit quantization serves as a viable alternative to enhance response quality without sacrificing throughput.}
    \item \new{\textit{Model Architecture and Sensitivity to Quantization}: Most models maintain performance comparable to the F16 baseline across various bit-widths, with significant degradation occurring only at extremely low levels, such as 3-bit. However, there is an exception—higher bit-widths do not necessarily improve accuracy for some models on certain tasks. For example, on TruthfulQA~\cite{lin-etal-2022-truthfulqa}, Llama$2$-$7$B and Gemma$3$-$1$B experience accuracy declines %
    when transitioning from Q$3$\_K\_M to Q$4$\_K\_M.}
    \end{itemize}

\textbf{Response to Q3:}
    \begin{itemize}
        \item \textit{\revision{DVFS Strategy}}: The DVFS strategy could be more aggressive. %
        As inference on mobile devices is typically intermittent, it is acceptable to increase the CPU frequency to speed up for single-round inference, as users are particularly sensitive to the first token generation latency.
        \revision{In contrast, GPUs on non-Apple devices tend to operate at a relatively fixed frequency (not the maximum). Therefore, there is still room for further acceleration when performing inference on GPUs.}
        \item \textit{\revision{Computing Thread}}: Given the ``big.LITTLE" architecture of mobile CPUs, in most cases, setting the number of threads equal to the number of big cores (prime and performance cores) can achieve the optimal performance. Adding more cores (efficiency cores) may degrade the performance.
        \revision{When running in parallel with other tasks, the optimal configuration should vary dynamically, depending on both the real-time workload and the nature of the co-running tasks.}
    \end{itemize}
\textbf{Response to Q4:}
    \begin{itemize}
        \item \textit{Instruction speedup}: Specialized machine instructions, such as smmla and sdot in Arm architecture, can significantly improve the performance. By carefully rearranging the weight matrix, these instructions can yield up to a \new{6× speed improvement for Llama$2$-$7$B.}
        \item \textit{Low GPU utilization on non-Apple SoC}: While LLMs are typically considered to be resource-intensive, the current implementation does not fully explore the potential of most mobile processors. It only utilizes $5$\% $-$ $20$\% of the arithmetic units on mobile GPUs in the prefill stage.
        \item \textit{Gap between Adreno and Mali GPUs}: Due to unaligned software implementations, processors with superior hardware specifications may not necessarily exhibit better performance (e.g., Mali-G$720$ Immortalis MP$12$ vs Adreno $750$). In addition, the Adreno GPUs consistently outperform the Mali GPUs in overall performance.
        \item \textit{\revision{Memory bottleneck:}} \revision{Due to the memory-bound nature of decoding, the decode speed does not vary significantly across top-tier devices. In contrast, during the prefill stage, suboptimal data layouts and memory access patterns can substantially reduce computational efficiency.}
    \end{itemize}

We summarize our contributions as follows:
\begin{enumerate}
    \item We present a comprehensive measurement study of LLM inference on mobile platforms. We present not only metrics relevant to user experience, but also hardware and system characteristics that are critical to developers. 
    \item We provide a thorough analysis of how hardware attributes and system dynamics impact on-device LLM performance and highlights potential bottlenecks in mobile LLM applications. 
    \item Based on our measurement results, we propose several potential directions to accelerate on-device LLM inference.
\end{enumerate}

\section{Related Works}
\label{sec:related_works}
\subsection{Light-Weight LLMs}
The number of LLMs has increased significantly in recent years. Numerous studies have explored the potential performance limits of these models by leveraging scaling laws~\cite{kaplan2020scaling} to increase their size. However, given the severe constraints of mobile platforms \new{and the high deployment costs~\cite{xiong2024search}~\cite{li2025towards}}, it is crucial to compress these models to ensure that they are lightweight enough for deployment on edge devices.
Current methods for compressing large models usually fall into the \new{two principal approaches: algorithmic-level and infrastructure-level~\cite{li2025towards}. The algorithmic-level optimization includes pruning~\cite{ma2023llm}\cite{xia2023sheared}, distillation~\cite{timiryasov2023baby}, low-order decomposition~\cite{xu2023tensorgpt}, and efficient attention implementation~\cite {yin2025dynamicsparseattentionmobile}~\cite{10.5555/3600270.3601459}. The infrastructure-level optimizations are exemplified by quantization~\cite{frantar2022gptq}\cite{lin2023awq}}. Quantization is a widely used resource-efficient technique for compressing LLMs, reducing their memory and bandwidth requirements. The nature of quantization is a mapping from floating-point numbers to integers, especially the 4-bit integer quantization for mobile LLMs, which is recognized as the optimal compromise between model size and accuracy~\cite{li2024large}. For example, based on the second-order information for weight updating, GPTQ~\cite{frantar2022gptq} employs layer-wise quantization to minimize the increase in global loss. K-quant~\cite{gg2024kquant} is a group-quantization method that divides the weight matrix into multiple $16\times8$ blocks, performing min-max quantization within each block. Due to outliers and dequantization overhead, most of the quantization algorithms focus on weight quantization only. However, by treating outliers carefully and implementing effective kernels, SmoothQuant~\cite{xiao2023smoothquant} and OdesseyLLM~\cite{li2023speed} have tried weight-activation co-quantization to further lower the costs of LLM inference.

To optimize hardware resource utilization and accelerate inference, operator fusion and key-value cache are commonly used for on-device LLMs. Operator fusion~\cite{li2023speed} significantly improves computational efficiency by reducing the number of operations and eliminating the need to copy intermediate results between operations. Besides, as the inference process of generative LLM is autoregressive, implementing kv-cache can significantly reduce the computational cost by reusing key and value vectors across different positions. Although this increases memory requirements, the amount of cache space used is manageable due to single-batch inference and the relatively short context length typical of mobile applications. In addition, to address the quadratic cost of attention, MQA~\cite{shazeer2019fast} and flash-attention~\cite{dao2022flashattention} have optimized existing algorithms to accelerate inference by minimizing the computational overhead involved in attention calculations.

The researchers have attempted to develop lightweight LLM models with less than $10$ billion parameters and optimized them with these resource-efficient algorithms, which facilitate local LLM deployment. For example, Meta's Llama 2 7B and Llama 3 8B models~\cite{dubey2024llama}, after instruction fine-tuning, show effectiveness across various tasks. Built upon a similar block structure as Llama-2 but more lightweight, phi-3 models~\cite{phi} introduced by Microsoft including phi-3-mini with 3.8B parameter and phi-3-small with 7B parameter achieves a quality that seems on-par with GPT-3.5. Moreover, Google's Gemini Nano~\cite{gemini}, specifically designed for mobile devices, comes in 1.8B and 3.25B sizes, which have been integrated into Pixel 8 Pro and can perform text summarization and provide intelligent responses locally. 

\subsection{On-device Deployment of LLMs}
Traditional machine learning (ML) and deep learning (DL) models have established a rich and mature ecosystem for mobile platforms. For example, TFLite, a DL inference engine designed for mobile devices, offers a wide range of highly optimized operators. Additionally, some vendors such as Qualcomm have developed hardware-specific DL libraries, which enhance the efficiency of model deployment and inference. However, unlike convolutional models, which typically have only a few hundred MB of parameters, LLMs feature orders of magnitude more parameters, more complex architectures, and higher computational demands. \new{Although recent studies~\cite{lyu2025larger} have demonstrated the efficient deployment of Transformer-based models at the edge, models with only millions of parameters exhibit a significant gap compared to contemporary billion-parameter LLMs in terms of both performance and deployment complexity. This complexity requires the development of specialized engines or operators to reduce latency and memory management strategies to minimize memory costs, all specifically designed for efficient LLM inference.}
\new{}

There are some inference frameworks widely used including llama.cpp~\cite{llamacpp}, MLC LLM~\cite{mlc-llm}, mnn-llm~\cite{mnnllm}, mllm~\cite{mllm}, etc. Among them, llama.cpp and mnn-llm are mainly optimized for LLM inference on CPUs, while MLC LLM leverages the computational capabilities of GPUs. \new{Despite the availability of dedicated AI accelerators such as NPUs, APUs, and TPUs in the latest generation of mobile SoCs~\cite{ignatov2019ai}, developing on these accelerators presents challenges due to the closed-source nature of vendor-specific SDKs. }%

In addition to those open-source LLM inference engines, mobile SoC vendors and manufacturers have also realized the importance of on-device LLM inference. Qualcomm, the vendor of Snapdragon SoCs, asserts that it can accelerate \revision{ Llama-2 7B using the NPU on Snapdragon 8 Gen3~\cite{QualcommNPU2}. } Similarly, MediaTek has announced optimizations for Llama 2 Generative AI applications using its APU on the Dimensity 9300 plus~\cite{mediatekAPU}. However, both solutions remain closed-source and have yet to be validated through real-world applications.

\subsection{Existing Measurement Studies}
Several studies have focused on the measurement and evaluation of mobile-optimized LLMs. \cite{laskaridis2024melting} and \cite{li2024large} both develop comprehensive benchmark suites to gain insights into various performance metrics, including latency, memory footprint, and throughput, as well as the impact of model size and quantization precision on accuracy. \cite{laskaridis2024melting} considers a range of edge devices, from smartphones to Nvidia Jetson. In addition to throughput, it also investigates device power consumption over time. \cite{li2024large} primarily compares the performance of 22 lightweight models, offering a detailed analysis on model structure and operators.

However, none of these studies provide in-depth analysis about the impact of mobile hardware during inference, such as accelerator utilization, memory bandwidth, and \revision{the influence of DVFS on both CPU and GPU, heat buildup during multiple runs, parallel execution with other tasks}. Although \cite{li2024large} touches on the impact of SoCs, it only examines the macro differences of SoCs from a single vendor. In contrast, this study involves testing several SoCs from multiple vendors to gain a comprehensive understanding of how mainstream mobile processors support LLMs. This will help us identify hardware bottlenecks and potential future upgrades necessary for optimizing LLM performance on mobile devices.

\begin{table*}[!t]
  \small
  \caption{Testing devices and their hardware specifications}
  \label{tab:params}
  \resizebox{\textwidth}{!}{\begin{tabular}{llccccccccl}
    \toprule
    Device & SoC & CPU & GPU & Avail./Total RAM & Type & OS & Release\\
    \midrule
    Xiaomi14 Pro&Snapdragon 8 Gen3&Cortex-X4/A720/720/A520&Adreno 750&12GB/16GB&top-tier& HyperOS & 2023.10 \\\hline
    Xiaomi Pad6 Pro	&Snapdragon 8+ Gen1	&Cortex-X2/710/510&Adreno 730&12GB/16GB& high-tier& Android13 & 2022.05 \\\hline
    Huawei Matepad11 Pro&Snapdragon 870&Cortex-A77/A77/A55&Adreno 650&5.5GB/8GB&mid-tier& Harmony 4& \rev{2021.01}\\\hline
    Vivo Pad3 Pro&Dimensity 9300&Cortex-X4/X4/A720&\makecell{Mali-G720 \\ Immortalis MP12}&10GB/16GB&top-tier& Android 14 & 2023.11\\\hline
    Huawei Matepad12.6 Pro&Kirin 9000E&Cortex-A77/A77/A55&Mali-G78 MP22&8.5GB/12GB&high-tier& Harmony 4& 2020.10\\\hline
    Huawei Nova7&Kirin 985&Cortex-A76/A76/A55&Mali-G77 MP8&4GB/8GB&mid-tier& Harmony 4& \rev{2020.04}\\\hline
    \revision{iPad Pro 11-inch (3rd generation)} & \revision{Apple M1} & \revision{Unknown (8-Core)} & \revision{Unknown (8-Core)}  & \revision{5.5GB/8GB} & \revision{top-tier} & \revision{iPadOS 18.2.1} & \revision{2021.04}\\
    \bottomrule
  \end{tabular}}
\end{table*}

\section{Experimental Setup and Methodology}
\label{sec:setup}
\subsection{Experimental Setup}

\textbf{Testing Devices.} Table \ref{tab:params} shows the devices used in our measurement. \revision{There are $7$ mobile devices from four SoC vendors: Snapdragon, MediaTek, Hisilicon and Apple.} 
All CPUs feature eight Arm Cortex cores but differ in implementation details such as big.LITTLE configurations and cache sizes. It is noteworthy that for \revision{non-iOS devices}, Snapdragon 8 Gen3 and Dimensity 9300 represent the most advanced SoCs available on mobile devices. We also test devices ranging from high-tier to mid-tier to capture a broad spectrum of hardware heterogeneity. \revision{Specifically, the top-tier devices represent the current upper bound for on-device LLM inference; high-tier devices capture the gains from recent hardware improvements, and mid-tier devices test the feasibility of running high-accuracy models on mainstream phones.} In terms of GPU, the vast majority of mobile GPUs are from Adreno (Qualcomm), Mali (Arm). Thus, we designate Snapdragon SoCs as representatives of Adreno GPUs and the rest as representatives of Mali GPUs. \revision{For iOS device, the iPad selected is equipped with a laptop-level SoC. Because iOS imposes strict per-app memory limits, lower-tier Apple devices cannot feasibly host our target LLM, so we only include a top-tier device.} 

\vspace{0.05in} \noindent \textbf{\new{Testing Model.}} 
\new{We evaluate six representative LLMs from the Llama, Qwen, and Gemma families, with parameter sizes ranging from $1$B to $7$B (Table \ref{table:model_specifications}). This selection represents a range of mobile application scenarios, from lightweight models for tasks like message summarization to larger models designed for autonomous agents. Since most mobile devices come with $8$GB to $16$GB of RAM, with at least $4$GB available after accounting for the operating system, $7$B-model is the upper limit for hardware compatibility. By default, we employ Llama$2$-$7$B with $4$-bit quantization as the evalutaion model.}

\vspace{0.05in} \noindent\textbf{Frameworks and Engines.} In this paper, we would like to reveal the performance of LLM on a variety of different accelerators. To achieve this, we select two representative frameworks: llama.cpp for CPU deployments, and MLC LLM for \revision{general} GPU deployments, to assess the inference capabilities of general accelerators. Additionally, llama.cpp supports GPU for specific SoCs, such as certain Snapdragon SoCs and Apple SoCs. %

\begin{itemize}[leftmargin=*]

    \item llama.cpp~\cite{llamacpp} is a pure C language project based on the GGUF machine learning tensor library. The model weights are quantized into lower accuracy by K-quant method and stored as binary files in GGUF format. After compiling the code into a specific back-end executable file, users can try different models for inference and only need to download the corresponding model weight. There is no need to rely on other environments. 
    \item MLC LLM~\cite{mlc-llm} is a high-performance LLM deployment engine. Built on the TVM machine learning compiler, MLC LLM is designed to harness the power of mobile GPUs. Leveraging TVM, MLC LLM can automatically generate and optimize operators for a specific model and backend. These operators are then loaded and executed via the TVM Runtime. For Android-based GPUs, MLC LLM utilizes the OpenCL backend to ensure efficient execution. %
\end{itemize}

\vspace{0.05in} \noindent \textbf{Prompts and Outputs.} To evaluate LLM performance in real-world scenarios, we use two types of prompts: a short prompt with $64$ tokens and a long prompt with $512$ tokens. The short prompt simulates typical user queries, such as asking for the meaning of a phrase, while the long prompt is designed for scenarios requiring more context, such as summarizing a news article. For generation, llama.cpp has fixed output lengths of $128$ tokens for $64$-token prompts and $256$ tokens for $512$-token prompts. In contrast, MLC LLM supports variable output lengths, allowing for generation up to \revision{$512$} tokens.

\vspace{0.05in} \noindent \textbf{Measurement toolkits.} In order to collect fine-grained hardware metrics such as real-time CPU frequency, utilization, and memory bandwidth \revision{on non-Apple devices}, we choose Perfetto~\cite{perfetto}, Snapdragon Profiler~\cite{snapdragon} and Arm streamline~\cite{streamline} as our primary tools for monitoring hardware dynamics during inference. Perfetto offers comprehensive system-wide performance traces from Android devices, including data from the kernel scheduler, userspace instrumentation, and other sources. For CPU monitoring, Perfetto captures frequency, utilization, and event scheduling for each CPU core. Snapdragon Profiler and Arm Streamline are vendor-specific profilers used to track GPU behavior, including arithmetic and load/store unit utilization. Specifically, Snapdragon Profiler provides detailed timelines of OpenCL kernel executions, which is valuable for operator-level analysis. \revision{As for the Apple device, we use Xcode~\cite{Xcode} to monitor the memory allocation, real-time bandwidth and GPU utilization.}

\subsection{Workflow and Methodology}

\begin{figure}[!t]
  \centering
  \includegraphics[width=\linewidth]{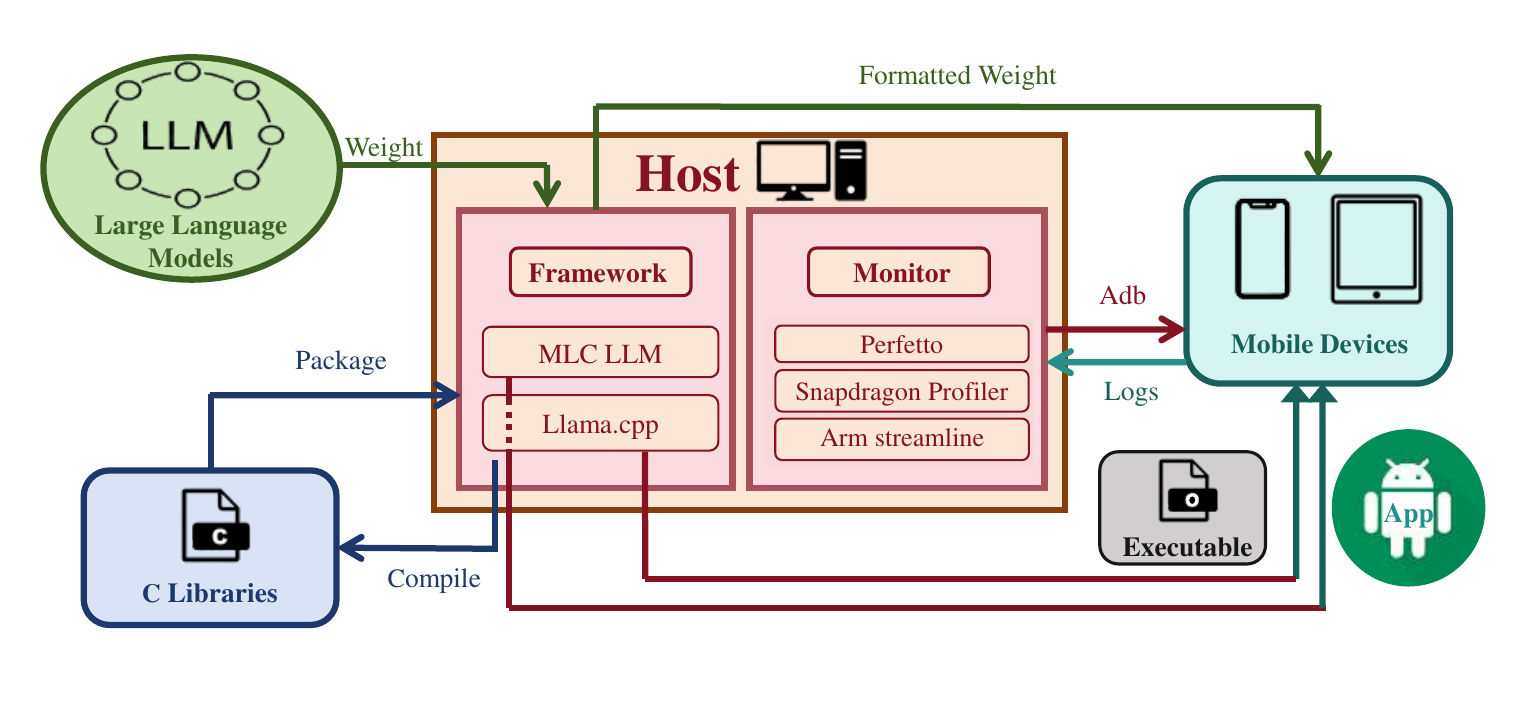}
  \caption{Measurement workflow.}
  \label{fig:workflow}
\end{figure}

\vspace{0.05in} \noindent \textbf{Workflow.} Figure~\ref{fig:workflow} shows the overall workflow how we deploy and evaluate models on mobile devices. The entire workflow is divided into four steps:
\begin{itemize}[leftmargin=*]
    \item \textbf{Step 1}: Compile the model into a specific model library or binary executable file that comprises the inference logic.
    \item \textbf{Step 2}: Transfer the library or executable file, model weights, and other necessary files to the test device and complete the deployment. 
    \item \textbf{Step 3}: Run the executable file or launch the application via ADB\cite{adb} and monitor hardware metrics with profilers\cite{perfetto}\cite{streamline}\cite{snapdragon} \revision{\cite{Xcode}} during inference. 
    \item \textbf{Step 4}: Collect results such as inference latency, CPU frequency fluctuations, and other key metrics.
\end{itemize}

For llama.cpp, we build the model library and compile the source code into an executable file on each device individually, using CMake and the Android NDK. All the compile options are kept the same by default. When necessary, we adjust the compile options to enable specialized instructions to ensure optimal performance (Section~\ref{subsec:instruction}). Since it relies on basic C libraries, no additional environment setup is required. After transferring model weights in GGUF format to the device, inference can be performed easily using the ADB command-line tool. For MLC LLM, there is a native application which TVM runtime and necessary libraries are packed in. The APK can be installed on the device, allowing interaction with the LLM through a graphical interface. MLC LLM cross-compiles the LLM models for the mobile platform, and the runtime version (including tvm and java) is the same on all devices.

\vspace{0.05in} \noindent \textbf{Measurement Methodology.} \label{method}
\new{For CPU-based inference, in order to minimize the effects of DVFS and ensure reliable results, we conduct two rounds of testing. In every round, each test (one round of inference) is repeated five times, with a 10-second interval between repetitions. After the first round, the device is rebooted and allowed to rest for 10 minutes to prevent overheating. For GPU-based inference, since the GPU frequency is typically stable, we perform five tests with a minimum 10-minute rest period between each test. The results are then averaged to provide a reliable measure of performance.}

\begin{table}[!tb]
  \centering
  \small
  \caption{\new{LLM model specifications}}
  \resizebox{0.46\textwidth}{!}{\begin{tabular}{ccc}
    \toprule
    Model Name & Model Size &  Detailed Version (Huggingface)  \\
    \midrule
    Llama$3.2$-$1$B &  $1.23$B & meta-llama/Llama-$3.2$-$1$B-Instruct \\
    Llama$3.2$-$3$B &  $3.21$B & meta-llama/Llama-$3.2$-$3$B-Instruct \\
    Llama$2$-$7$B &  $7$B & meta-llama/Llama-$2$-$7$b-chat-hf \\
    \midrule
    Qwen$3$-$1.7$B &  $1.7$B & Qwen/Qwen$3$-$1.7$B \\
    Qwen$3$-$4$B &  $4.0$B & Qwen/Qwen$3$-$4$B-Instruct-$2507$ \\
    \midrule
    Gemma$3$-$1$B &  $1.0$B & google/gemma-$3$-$1$b-it \\
    \bottomrule
  \end{tabular}}
  \label{table:model_specifications}
\end{table}

\section{Performance: Users' Perspectives}
\label{sec:users}

In this section, we present performance metrics that affect user experience, including token throughput, memory footprint, and energy consumption. We specifically compare the key performance metrics across %
CPUs and GPUs. %
\subsection{Token Throughput and Memory Footprints}
\subsubsection{\new{Performance on CPUs}}
\label{subsubsec:cpu}

\begin{figure*}[!tb]
\centering
	\subfloat[\new{Inference performance for (64, 128) prompts (with 8 threads)}]{\includegraphics[width = \linewidth]{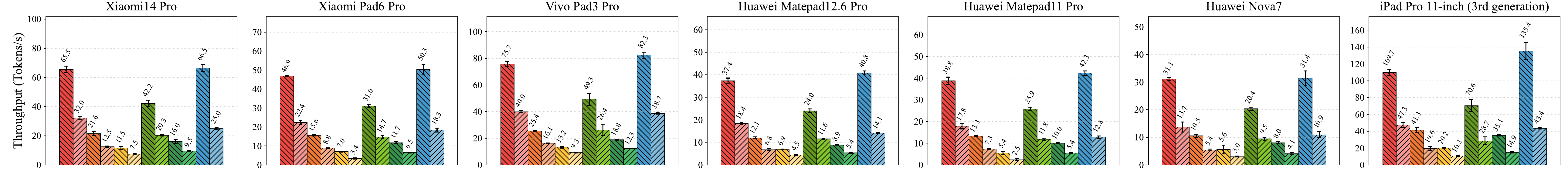}}
	\hfill
	\subfloat[\new{Inference performance for (512, 256) prompts (with 8 threads)}]{\includegraphics[width = \linewidth]{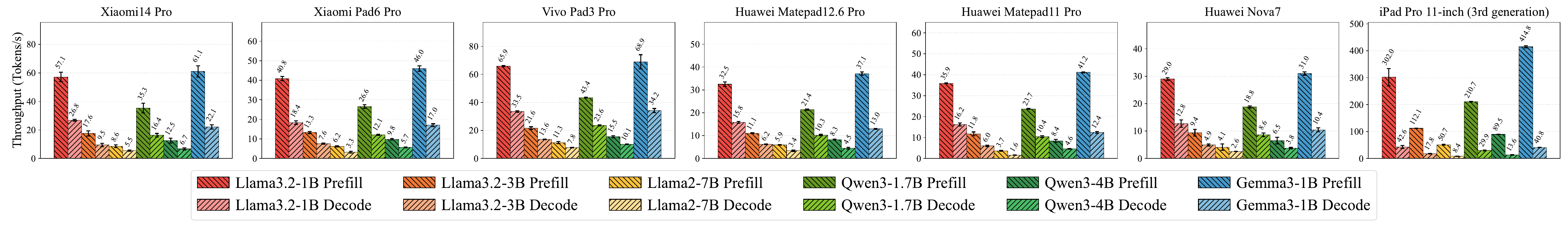}} 
\caption{\new{Inference performance on CPU (llama.cpp)}}
\label{fig:llamp_throughput}
\end{figure*}

\new{Figure~\ref{fig:llamp_throughput} shows the performance of CPUs on different mobile devices, including prefill speed and decoding speed with llama.cpp. From the device perspective, there has been a consistent improvement in both prefill and decoding speeds with new-generation processors. Among the tested devices, the Apple M$1$ SoC exhibits the most superior performance as a laptop-grade processor. The Dimensity $9300$, while leading the non-Apple SoC category, trails the Apple M$1$ by approximately $10$\% to $20$\% during the decode stage. However, its exceptional performance significantly diminishes in the prefill stage, where it achieves $60$\% of the Apple M$1$’s throughput for 64-token prompts and only $20$\% for $512$-token prompts. The Snapdragon $8$ Gen $3$ follows as the third-best device, delivering roughly $80$\% of the throughput observed in the Dimensity $9300$. Additionally, a significant performance gap exists between Kirin SoCs and their Snapdragon and MediaTek counterparts in terms of LLM support.}
\new{From the model perspective, we draw the following observations:}

\vspace{0.05in} \noindent \new{\textbf{\textit{Linear Scaling of Computational Overhead}}: The computational cost exhibits a nearly linear relationship with increases in model size. Across different devices and configurations, the throughput ratios closely mirror the model size ratios within the same family. For instance, after applying $4$-bit quantization, the Qwen$3$-$4$B model, which is roughly twice the size of Qwen$3$-$1$B, demonstrates a prefill speed that is only $0.3\times$ to $0.5\times$ of the Qwen$3$-$1$B when tested on the same device and under the same configuration.}

\vspace{0.05in} \noindent \new{\textbf{\textit{Stage-Specific Sensitivity}}: The growth of model parameters has a more noticeable effect on prefill performance than on decoding, as the prefill stage is compute-bound. Larger models, with increased hidden dimensions and depth, result in greater computational load. For instance, with Llama3 models, the prefill performance decreases by $5$\% to $10$\% more than decoding when scaling from $1$B to $3$B. This trend persists across devices%
, with larger models consistently exhibiting a similar pattern of performance degradation compared to their smaller counterparts. %
This pattern also holds for Qwen$3$ models.}

\vspace{0.05in} \noindent \new{\textit{\textbf{Architectural Influence on Throughput}}: Model architecture plays a critical role in determining both parameter distribution and execution efficiency. This is particularly evident when comparing models with nearly identical parameter size, such as Llama$3.2$-$1$B and Gemma$3$-$1$B. Despite both models occupying approximately $760$MB post-quantization, their performance metrics diverge across different inference stages. Specifically, Llama$3.2$-$1$B demonstrates slower prefill speeds but superior decoding performance compared to Gemma$3$-$1$B. These differences underscore the inherent architectural trade-offs between these two distinct inference phases.}

\vspace{0.05in} \noindent \new{\textit{\textbf{Sequence Length and Hardware Elasticity}}: Increasing the input sequence length intensifies both computational demand and memory pressure. On most non-Apple devices, extending the prompt length from 64 to 512 tokens results in a performance degradation exceeding $10\%$. In contrast, the Apple M1 architecture demonstrates remarkable elasticity: while decoding speed remains stable, prefill speed improves by a factor of $2.5$ to $3$ with longer prompts, highlighting significant potential for handling complex computations.}

\begin{figure*}[!tb]
  \centering
  \includegraphics[width=\linewidth]{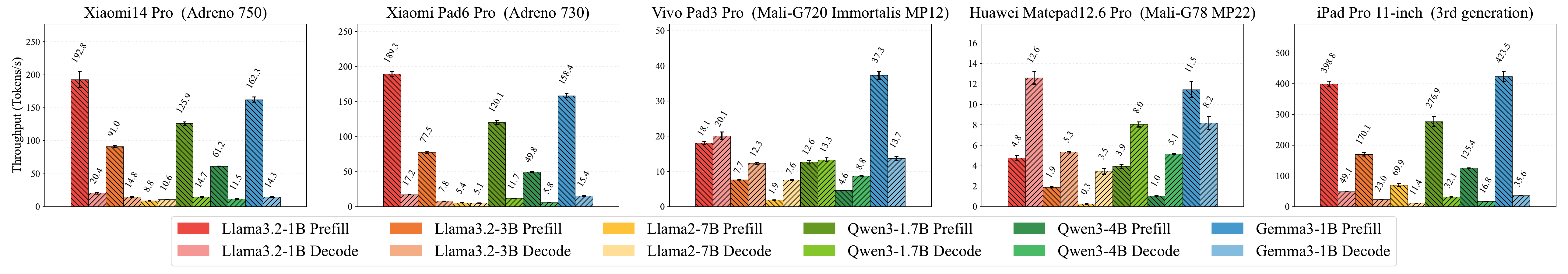}
  \caption{\new{Inference performance (MLC LLM) on GPU with 64-token prompt}}
  \label{fig:gpu_throughput}
\end{figure*}

In addition, memory footprint is also a crucial metric. The memory usage closely corresponds to the size of the model weights. On all six devices, memory usage remains consistently around $3.8$GB when running Llama$2$-$7$B \new{and below $1$GB} when running Gemma$3$-$1$B. Additionally, it shows minimal variation during inference once the model weights are loaded into main memory from external storage. %
This consistency is due to the fact that llama.cpp applies for a fixed-size memory block during initialization, which remains constant once allocation is complete. The allocated memory is then segmented into various components for different variables, including model weights, KV-cache,  buffers, context and etc.

\begin{observation}
LLMs running on mobile CPUs require only slightly additional memory beyond the model weights. Specifically, for a $4$-bit quantized lightweight LLM with $7$B parameters, the maximum memory usage is no more than $4$GB. Moreover, the parameters of many popular lightweight LLMs range from $1$B to $5$B, which further reduces memory requirement. Thus, as long as a mobile device has more than $4$GB available RAM, it is sufficient to run these models locally.
\end{observation}

\subsubsection{\new{Performance on GPUs}}

Unlike CPUs, which are all from Arm, GPUs across vendors may have different architectures. Thus, it is challenging for developers to optimize for different GPUs. Figure~\ref{fig:gpu_throughput} illustrates the end-to-end latency \del{of MLC LLM} across \revision{five} high-tier GPUs from Arm Mali, Qualcomm Snapdragon, and \revision{Apple}.
\revision{It should be noted that Huawei MatePad 11 Pro and Huawei Nova 7 exhibit frequent crashes when running MLC LLM, which prevents the collection of reliable measurement results. Consequently, these two devices are excluded from the comparison.}
Among the tested GPUs, Mali GPUs, particularly the Mali-G78, exhibit significantly slow prefill speeds. 
\new{Consequently, for ease of comparison, we focus only on the performance with 64-token prompts. It should be noted that the test for Llama2-7B on the Mali-G78 was conducted using 32-token prompts due to crashes that frequently occurred with longer prompts.}

\vspace{0.05in} \noindent \new{\textbf{\textit{Hardware Perspective}}: The results highlight a significant performance gap between Apple SoCs and non-Apple SoCs. For models around $1$B, the throughput on the Apple M1 is approximately $2\times$ faster than on the Adreno $750$. As for non-Apple SoCs, the Adreno GPU consistently outperforms the Mali GPU in overall performance across different models. Despite Mali GPUs showing poor prefill performance, the Mali-G$720$ achieves a slight speedup in decoding compared to the older-generation Snapdragon SoC, i.e., the Adreno $730$.}

\new{For the two Mali GPUs, both prefill and decode speeds are slower than on the CPUs. The slowdown is especially significant in the prefill stage.
For models larger than $4$B, the prefill speed drops below $5$ tokens/s on the Mali-G$720$, despite it having the highest theoretical computational capability among all non-Apple devices (see Table~\ref{tlb:gpu_specifications}). This phenomenon and its underlying causes will be further explored in Section~\ref{subsubsec:gpu_utilization}.}

\vspace{0.05in} \noindent \new{\textbf{\textit{Model Perspective}:} From the model perspective, we observe that for both Apple and Snapdragon GPUs, the prefill speed improves several times over %
CPU on the same device. For instance, on the Apple M$1$, the prefill speed of all models is $3$ to $4\times$ faster. However, the decode speed on the GPU is quite similar to that on the CPU, indicating that while the GPU offers a clear advantage in computation and can greatly accelerate prefill, the decode stage remains the bottleneck that limits overall throughput.}

\new{In the prefill stage, the Llama$3$ models show the highest improvement on both Apple and Snapdragon devices. For example, on the Snapdragon $8$+ Gen$1$, Llama$3.2$-$3$B achieves a $4\times$ improvement in prefill speed. For the $1$B models from Gemma$3$ and Llama$3.2$, despite the similar model sizes, the prefill acceleration for Llama$3.2$-$1$B is higher than Gemma$3$-$1$B on the same device. This suggests that the architecture of Llama$3.2$ is more efficient for parallel computing.}

\new{Additionally, memory usage across the GPUs is nearly identical, around $4.2$GB when running Llama$2$-$7$B, with a slight increase to $4.4$GB observed on the Vivo Pad$3$ Pro. However, as the prompt length and prefill chunk size increase, memory usage grows accordingly, showing higher memory demand compared to inference on CPU.}

\begin{observation}
\new{Although GPUs do not exhibit a clear advantage for large models (Llama$2$-$7$B), they show a significant improvement in prefill for smaller models (\eg, $1$B, $3$B). It is clear that hybrid execution, such as performing prefill with the GPU and decoding with the CPU, is an effective strategy. Furthermore, from the suboptimal performance observed on Mali GPUs, %
GPU acceleration often requires device-specific tuning to overcome architectural disparities and achieve competitive efficiency.}

\end{observation}

\subsection{\revision{Parallel with Background Tasks (non-AI tasks)}}
\revision{
In real-world scenarios, users often run inference alongside other processes, which can create a gap between peak and actual performance. To gain a deeper understanding of performance under typical conditions, we expand our evaluation of LLM inference by incorporating concurrent background workloads.}

\revision{The experimental configuration remains unchanged from the one described in Section~\ref{method}, with the exception that a music application is launched after rebooting the device and kept running in the background during the inference process. All data are collected on the Xiaomi $14$ Pro using llama.cpp. Given that the additional CPU utilization introduced by the music application is below $10\%$, we limit the number of parallel threads to $6$ and $8$ when running llama.cpp}.

\begin{figure}[tb]
    \centering
    \begin{subfigure}[b]{\linewidth}
        \centering
        \includegraphics[width=0.9\linewidth]{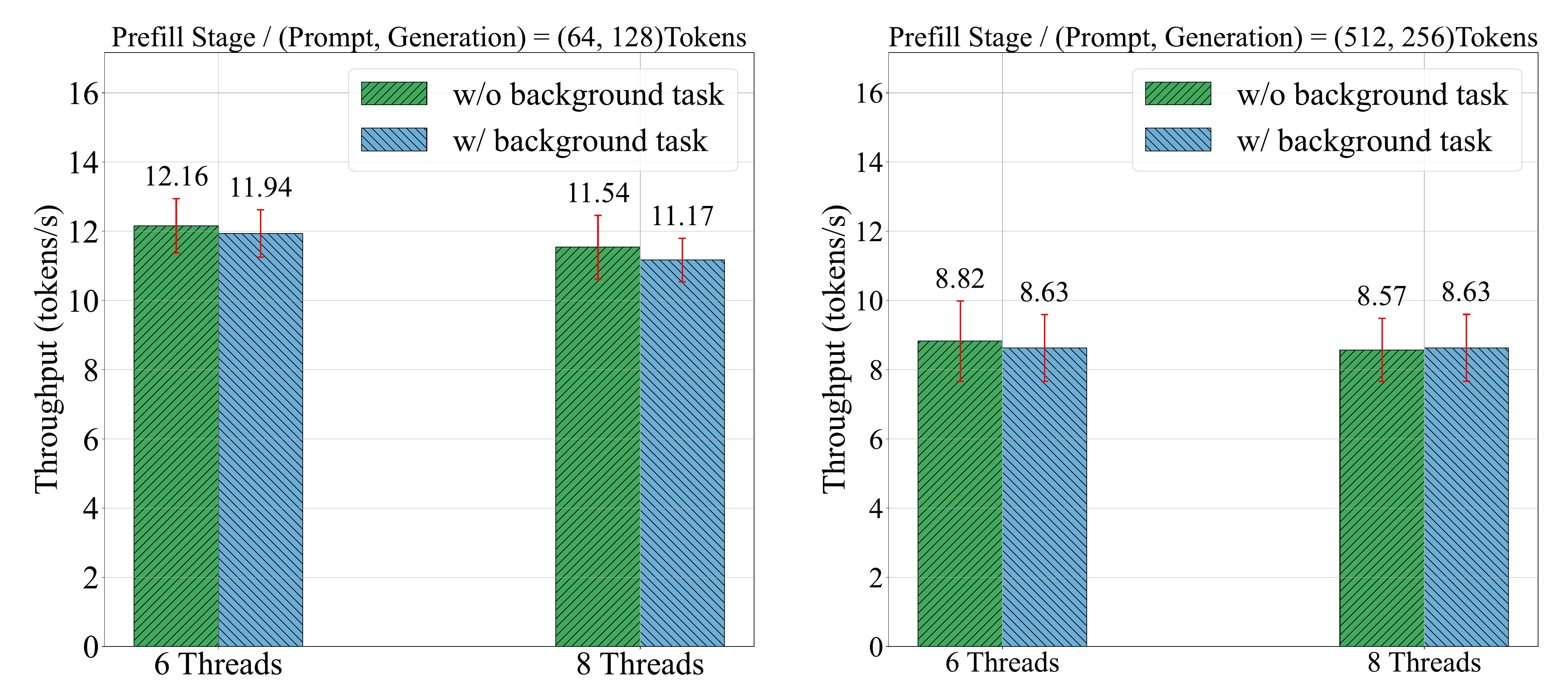}
        \caption{\revision{Throughput in Prefill Stage}}
        \label{fig:backtask-a}
    \end{subfigure}
    \\
    \vspace{2mm}
    \begin{subfigure}[b]{\linewidth}
        \centering
        \includegraphics[width=0.88\linewidth]{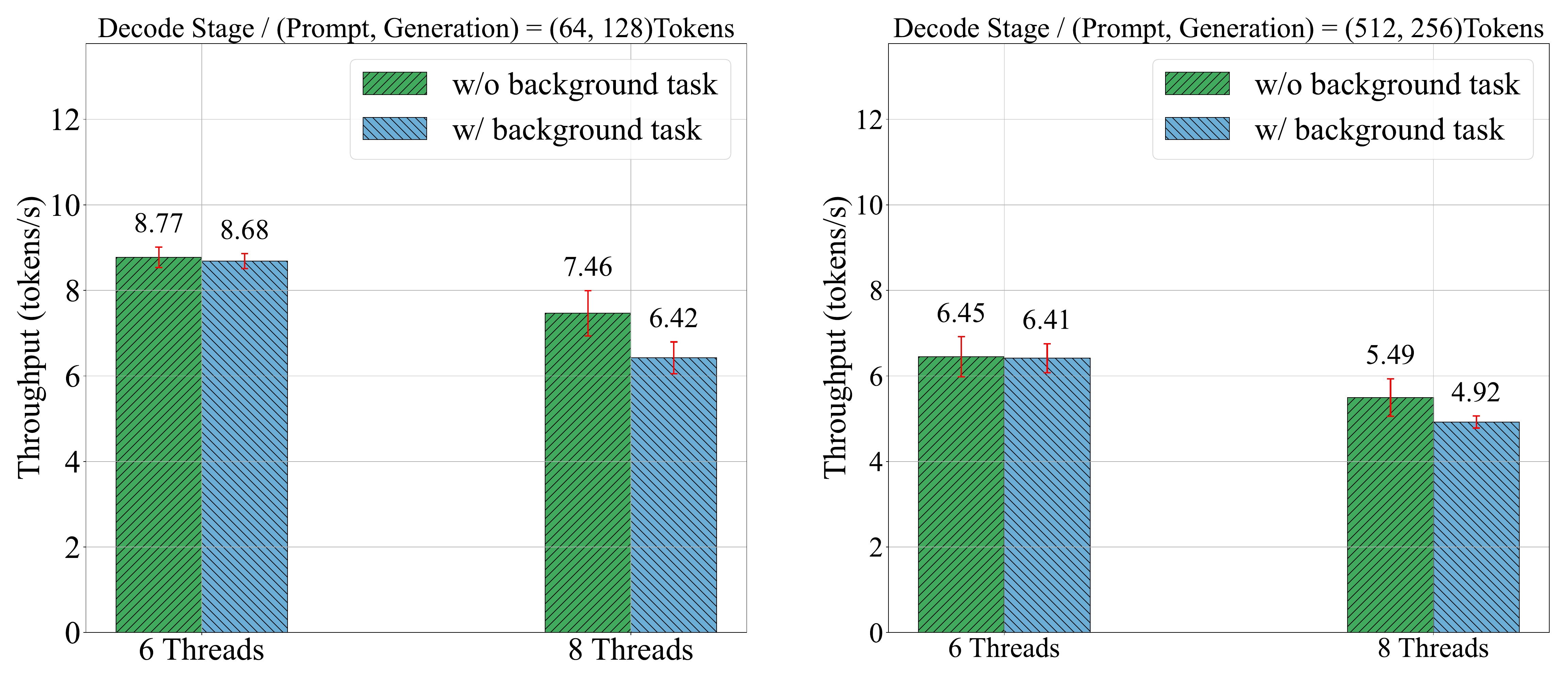}
        \caption{\revision{Throughput in Decode Stage}}
        \label{fig:backtask-b}
    \end{subfigure}
    \caption{\revision{Performance of Llama-2-7B with Background Task (Music App) on Xiaomi14 Pro.}}
    \label{fig:back_throughput}
\end{figure}

\revision{In Figure~\ref{fig:back_throughput}, for the prefill stage, background tasks cause only a minor performance drop, with differences remaining below $5\%$. A similar effect is observed during the decode stage when inference is executed with $6$ threads. However, when all CPU cores are utilized (\ie, $8$ threads), decode throughput degrades more significantly under background load, with performance losses exceeding $10\%$. Notably, while the background music application leaves CPU utilization largely unaffected, it reduces available RAM by several hundred megabytes, which likely contributes to the observed slowdown. These results suggest that inference performance will degrade further as the number or intensity of concurrent background tasks increases.} %

\subsection{\revision{Parallel with AI Tasks}} \label{sec:parallel_ai}
\revision{
In addition to small workloads, there are scenarios where LLM inference must be executed alongside tasks that heavily consume compute or memory resources. For example, some image processing neural network models are deployed on mobile devices. It is important to evaluate whether the performance remains acceptable when these tasks run concurrently.}

\revision{The experimental configuration is the same as in Section~\ref{method} except that a YOLO-based application is launched after rebooting the device and kept running in the foreground during inference. During LLM inference, the YOLO-based application activates the camera and executes the YOLOv$11$n model for object detection. All experiments are conducted on the Xiaomi $14$ Pro with llama.cpp.}

\begin{figure}[t]
    \centering
    \begin{subfigure}[b]{\linewidth}
        \centering
        \includegraphics[width=0.9\linewidth]{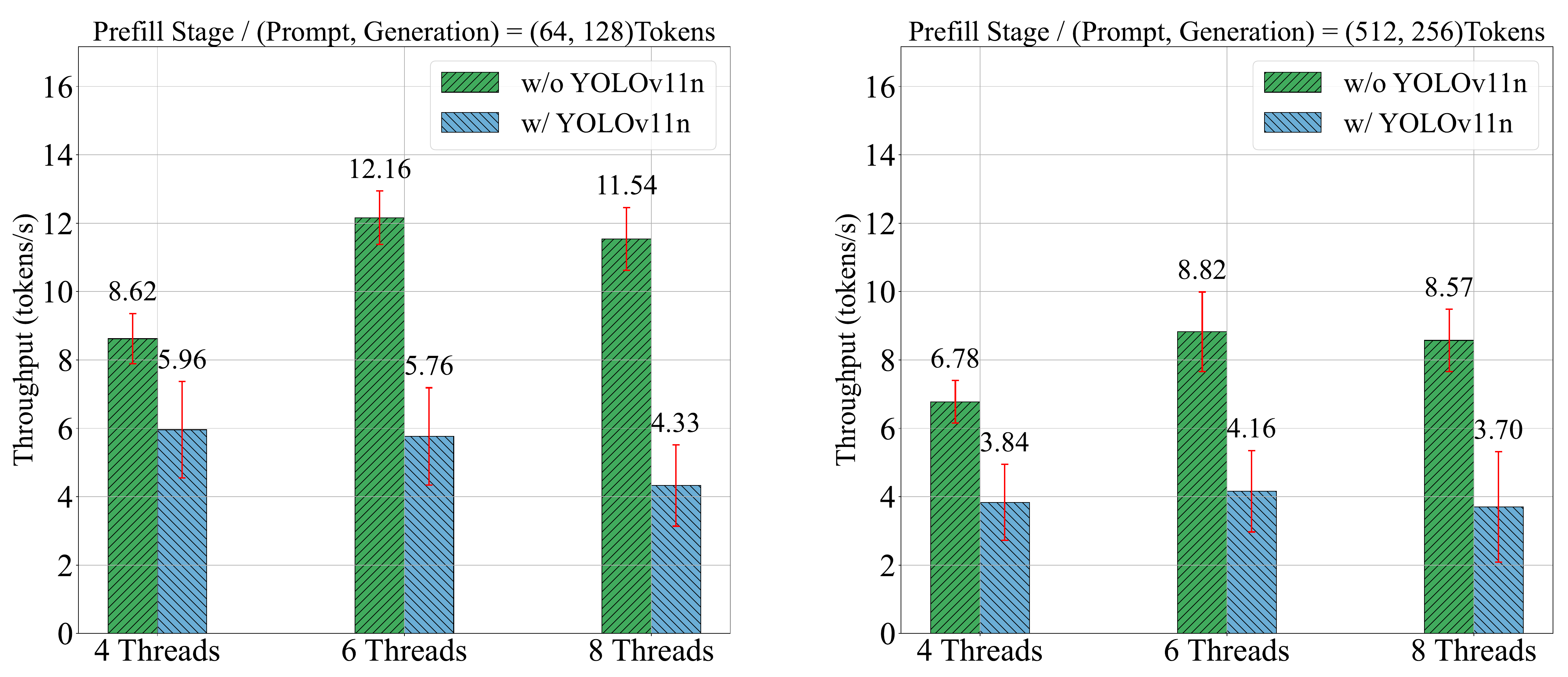}
        \caption{\revision{Throughput in Prefill Stage}}
        \label{fig:yolo-a}
    \end{subfigure}
    \\
    \vspace{1mm}
    \begin{subfigure}[b]{\linewidth}
        \centering
        \includegraphics[width=0.9\linewidth]{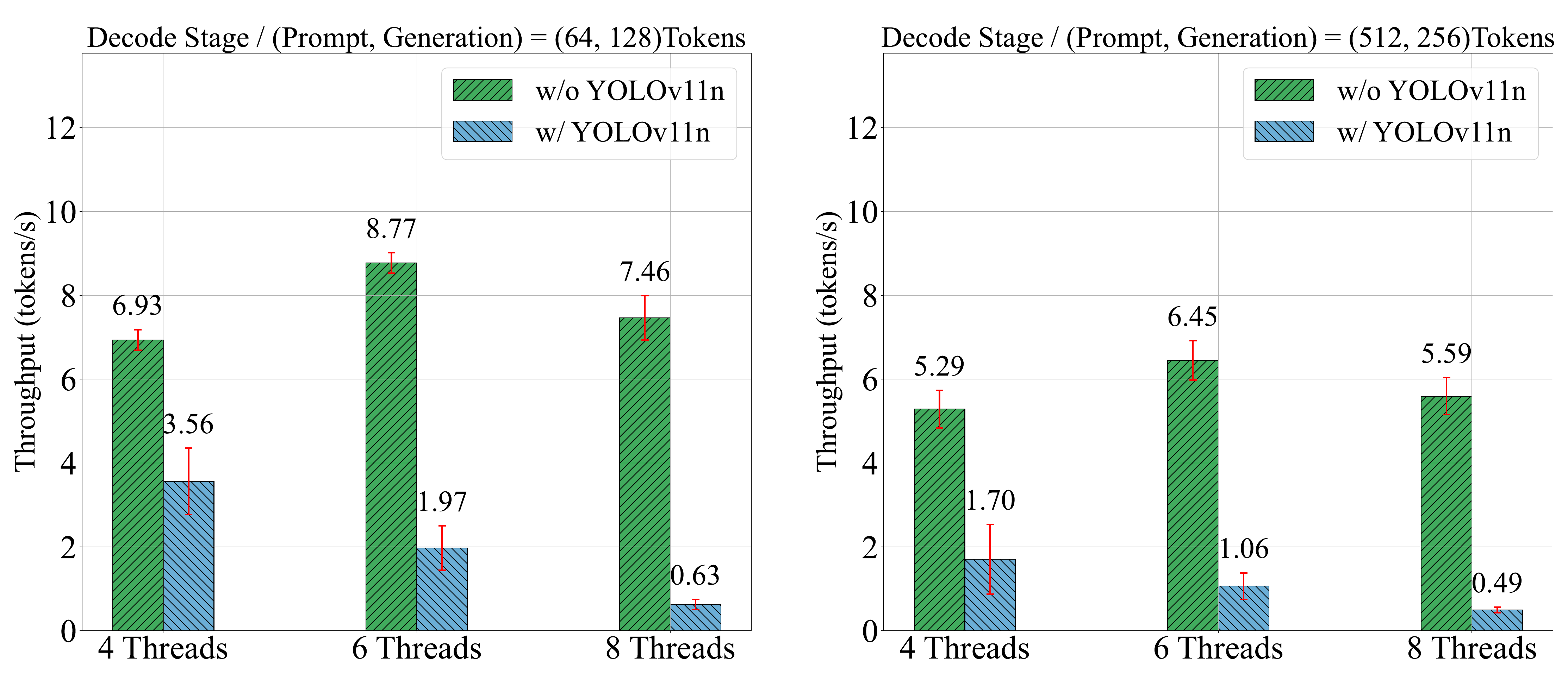}
        \caption{\revision{Throughput in Decode Stage}}
        \label{fig:yolo-b}
    \end{subfigure}
    \caption{\revision{Performance of Llama-2-7B with AI Task (YOLOv11n) on Xiaomi14 Pro.}}
    \label{fig:yolo_throughput}
\end{figure}

\revision{Figure~\ref{fig:yolo_throughput} shows that when executed concurrently with other AI tasks, inference performance declines progressively as the number of parallel threads in the LLM inference process increases. Without additional parallel tasks, the best performance is achieved at $6$ threads, where CPU utilization reaches approximately $680\%$. However, when the YOLO-based application is running (introducing an additional $\sim300\%$ CPU utilization), the optimal configuration shifts to $4$ threads, which yields the smallest relative degradation. Moreover, performance degradation in the decode stage becomes significantly more pronounced than in the prefill stage. The most critical case occurs at $8$ threads: prefill throughput drops by nearly $60\%$, while decode throughput collapses by more than $90\%$. In practice, this indicates that during the decode stage, inference with $8$ threads (equal to the number of CPU cores) becomes essentially unacceptable when YOLOv$11$n runs concurrently.}

\revision{
During execution with YOLOv$11$n, the battery temperature rose from $28^\circ\text{C}$ to $47^\circ\text{C}$ over five inference runs, which is substantially higher compared to LLM-only execution. This induces severe thermal throttling, forcing the prime core to operate at an extremely low frequency. Consequently, optimizing processor frequency to balance performance and energy consumption is crucial for concurrent execution.
}

\revision{
\begin{observation}
With YOLOv11n consuming over $300\%$ of CPU resources, the highest LLM inference performance is observed with fewer parallel threads. This suggests that the optimal thread count for practical scenarios is not fixed but instead depends on real-time CPU utilization. For AI workloads that demand substantial CPU capacity, parallel execution with LLMs presents significant challenges.
\end{observation}
}

\revision{
\begin{observation}
As the number of threads executing inference increases, the throughput of the decode stage experiences a pronounced decline, indicating that the memory bottleneck is the primary limitation when LLM inference is executed concurrently with other AI workloads. Under such multi-task scenarios, intensified memory read/write pressure and bandwidth contention significantly degrade inference performance, in some cases reducing it to the point of near unusability.
\end{observation}
}

\subsection{\new{Accuracy vs Latency}}

\begin{figure*}[!t]
  \centering
  \includegraphics[width=0.82\linewidth]{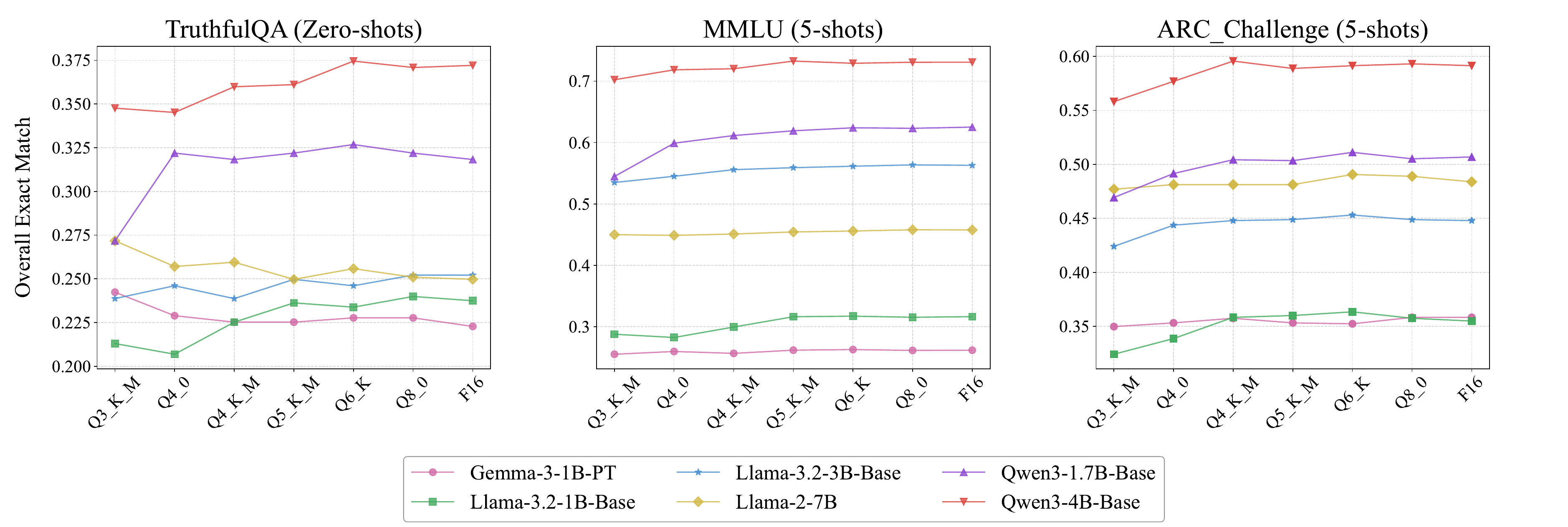}
  \caption{\new{Model accuracy on different tasks (TruthfulQA, MMLU, ARC)}}
  \label{fig:acc}
\end{figure*}

\new{We evaluate six pretrained models without fine-tuning across three benchmarks to assess various inference capabilities, including the MMLU~\cite{hendryckstest2021} and ARC~\cite{Clark2018ThinkYH} for general problem-solving, and TruthfulQA~\cite{lin-etal-2022-truthfulqa} for reliability and safety. As shown in Figure~\ref{fig:acc}, within a given model family, performance correlates positively with parameter scale. However, architectural differences among families introduce significant performance disparities. For instance, Qwen$3$ consistently outperforms the Llama and Gemma$3$ series. Notably, Qwen$3$-$1.7$B shows a substantial advantage over Llama$2$-$7$B, surpassing its MMLU accuracy by nearly $15$\% despite the latter having a larger parameter count. While Llama$3.2$-$1$B and Gemma$3$-$1$B demonstrate nearly identical accuracy in line with their similar scales, they are substantially outperformed by Qwen$3$-$1.7$B.}

\new{Regarding quantization schemes, most models maintain performance comparable to the F$16$ baseline across various bit-widths, with significant degradation occurring only at extremely low levels such as $3$-bit. Interestingly, transitioning from Q$3$\_K\_M to Q$4$\_K\_M on TruthfulQA yields divergent outcomes: while the accuracy of Qwen$3$-$1.7$B improves, Llama$2$-$7$B and Gemma$3$-$1$B experience declines exceeding $1$\%. These findings suggest that higher bit-widths do not necessarily enhance performance, underscoring the necessity of post-quantization evaluation to identify the optimal configuration for specific tasks.}

\begin{figure*}[!t]
\centering
	\subfloat[\new{Instruction acceleration on Xiamo14 Pro (with 6 threads)}]{\includegraphics[width = 1.0\linewidth]{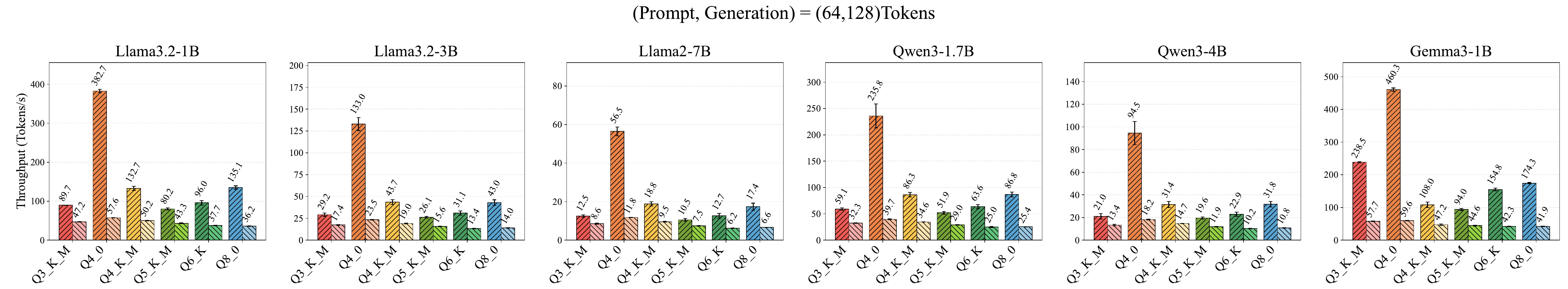}}
	\hfill
	\subfloat[\new{Naive inference on Huawei Pad 12.6 Pro (with 8 threads)}]{\includegraphics[width = 1.0\linewidth]{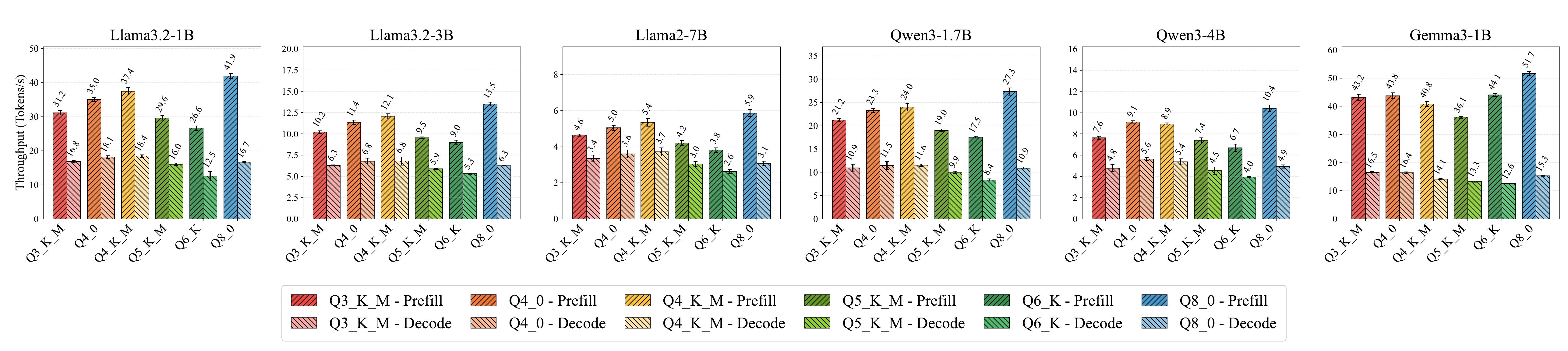}} 
    
\caption{\new{Inference throughput for different quantization types on CPU (llama.cpp)}}
\label{fig:throughput_quan}
\end{figure*}

\new{From the efficiency perspective, as shown in Figure~\ref{fig:throughput_quan}, $4$-bit quantization strikes the optimal balance by achieving peak or near-peak speeds in both prefill and decoding stages while maintaining robust inference accuracy. Leveraging the advanced instruction optimizations (see Section V.A), Q$4$\_0 achieves a significantly higher throughput on the Xiaomi $14$ Pro compared to other quantization schemes. For small-scale LLMs, such as Gemma$3$-$1$B, $8$-bit quantization serves as a viable alternative to enhance response quality without sacrificing throughput. In the compute-bound prefill stage, both $4$-bit and $8$-bit schemes exhibit clear advantages as their bit-widths align with the standard 8-bit byte structure. Schemes that do not align with this byte structure necessitate additional bitwise operations for dequantization, thereby increasing computational overhead.  Conversely, in the memory-bound decoding stage, despite the smaller memory footprint of 3-bit quantization, 4-bit quantization generally offers slightly higher throughput than its 3-bit alternative.}

\begin{observation}
    \new{The architectural diversity of mobile hardware necessitates a hardware-aware deployment strategy rather than a universal quantization standard. For top-tier devices with instruction-level acceleration, the Q4\_0 scheme offers an optimal balance by maximizing execution speed while preserving accuracy. In contrast, for resource-constrained platforms, pairing a smaller model with 8-bit quantization provides a superior compromise between response quality and practical throughput. Ultimately, on-device benchmarking of quantization is crucial to align the model with specific hardware constraints and the requirements of the target task, optimizing overall performance.}
\end{observation}

\section{Performance: Developers' Perspectives}
\label{sec:developers}
In this section, we present results that developers care about, focusing on CPU/GPU utilization, DVFS and scheduling strategies. We also investigate the impact of different inference frameworks. We hope that these results can help developers identify bottlenecks in LLM inference and ultimately lead to improvements in system performance. 

\subsection{CPU Utilization for LLM inference}
While upgrading hardware can enhance LLM local inference performance, it is also crucial to assess whether we are fully utilizing the capabilities of existing hardware. To address this, we use specialized profilers to monitor and capture dynamic utilization of the CPU and GPU during inference. This allows us to explore the potential of current hardware and identify opportunities for further accelerating LLM inference.

\subsubsection{Multi-threads on CPU Cores}
\begin{table*}[t]
  \footnotesize
  \centering
  \caption{CPU Specifications}
  \label{tab:cpu_specifications}
  \resizebox{1\textwidth}{!}{
    \begin{tabular}{ccccc}
      \toprule
      SoC & Snapdragon $8$ Gen $3$ & Dimensity $9300$ & Snapdragon $8$+ Gen $1$ & Kirin $9000$E \\
      \midrule
      Prime &  $1 \times$ Cortex-X4 ($3.3$GHz) &  $1 \times$ Cortex-X4 ($3.25$GHz) & $1 \times$ Cortex-X2 ($3.2$GHz) & $1 \times$ Cortex-A77 ($3.13$GHz) \\
      \midrule
      Performance & \makecell{$2 \times$ Cortex-A720 ($3.15$GHz) \\ $3 \times$ Cortex-A720 ($2.96$GHz)} & \makecell{$3 \times$ Cortex-A720 ($2.85$GHz) \\ $4 \times$ Cortex-A720 ($2$GHz)} & \makecell{$3 \times$ Cortex-A710 ($2.75$GHz)} & \makecell{$3 \times$ Cortex-A77 ($2.54$GHz)} \\
      \midrule
      Efficiency & $2 \times$ Cortex-A520 ($2.27$GHz) & - & $4 \times$ Cortex-A510 ($2$GHz) & $4 \times$ Cortex-A55 ($2.05$GHz) \\
      \bottomrule
    \end{tabular}
  }
\end{table*}

Most popular mobile SoCs utilize the ``big.LITTLE" architecture for their CPUs, which balances performance with power efficiency. This configuration typically includes multiple cores organized into two distinct clusters: ``big" (prime and performance cores) and ``little" (efficiency cores), as illustrated in Table~\ref{tab:cpu_specifications}. \revision{The detailed CPU Specifications of Apple M1 are not publicly disclosed; it is known to comprise 4 performance cores and 4 efficiency cores.} While it is commonly assumed that high-load tasks are best handled by the ``big" cores for optimal performance, our tests reveal that the ideal core configuration varies across the two stages of LLM inference.
\begin{figure}[!tb]
  \centering
  \includegraphics[width=0.95\linewidth]{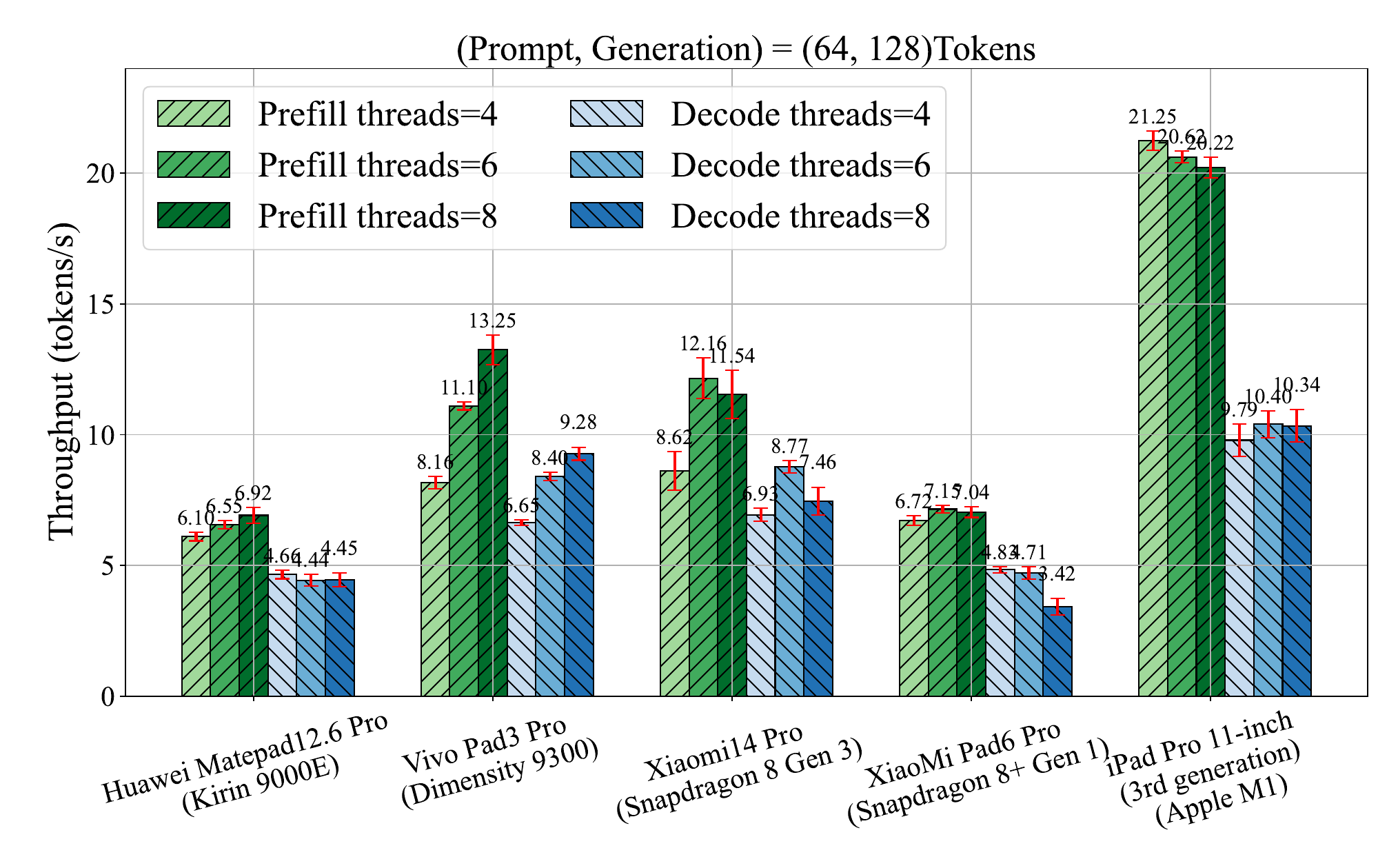}
  \caption{\revision{Throughput with multi-threads (llama.cpp with Llama$2$-$7$B)}}
  \label{fig:multi-threads}
\end{figure}

\begin{figure*}[!tb] %
    \centering
    \begin{minipage}[b]{0.49\textwidth}
       \centering
      \includegraphics[width=\linewidth]{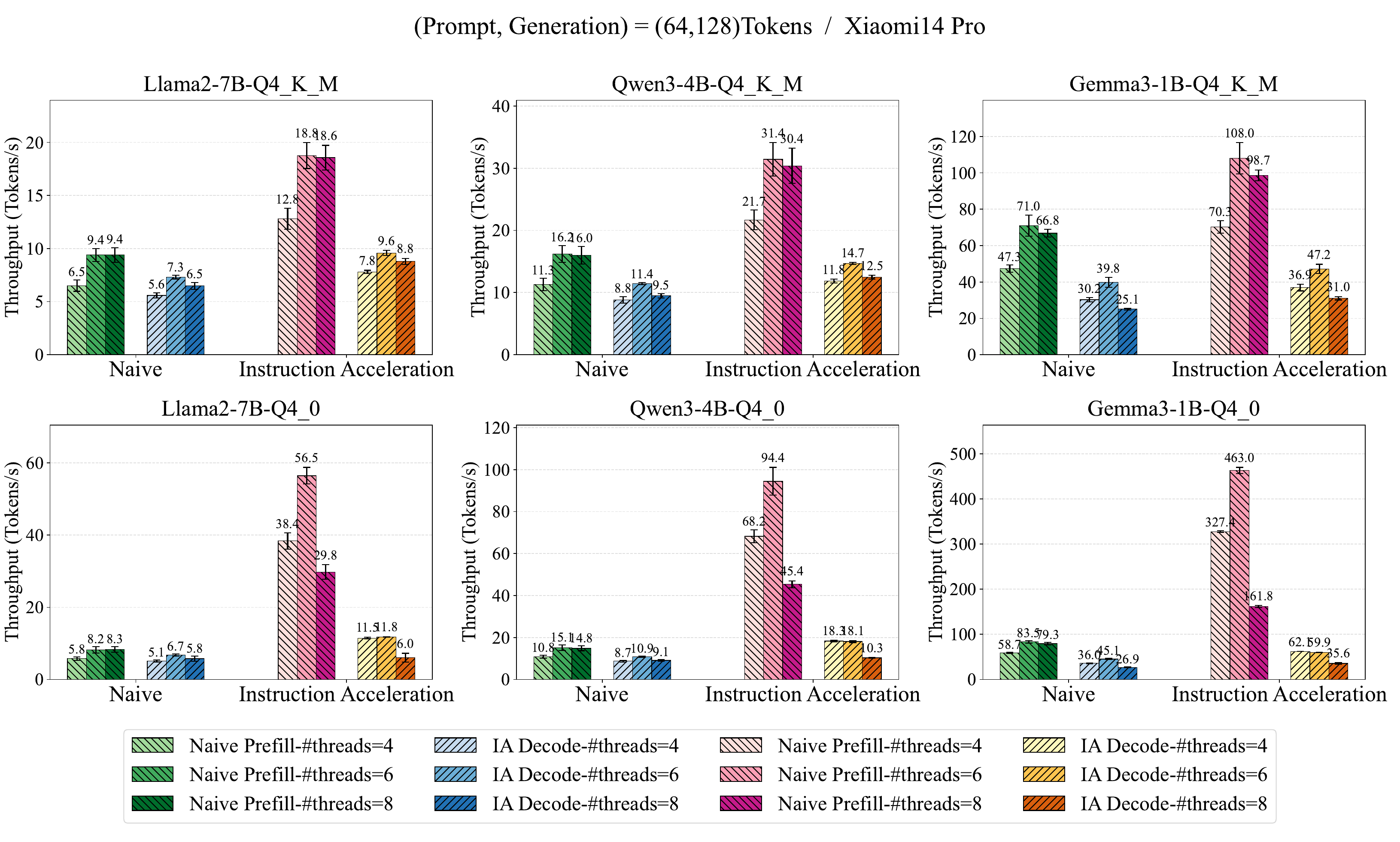}
      \subcaption[]{\new{Xiaomi14 Pro (Snapdragon 8 Gen3)}}
      \label{fig:i8_throughput_xm}
    \end{minipage}
    \begin{minipage}[b]{0.49\textwidth}
      \centering
      \includegraphics[width=\linewidth]{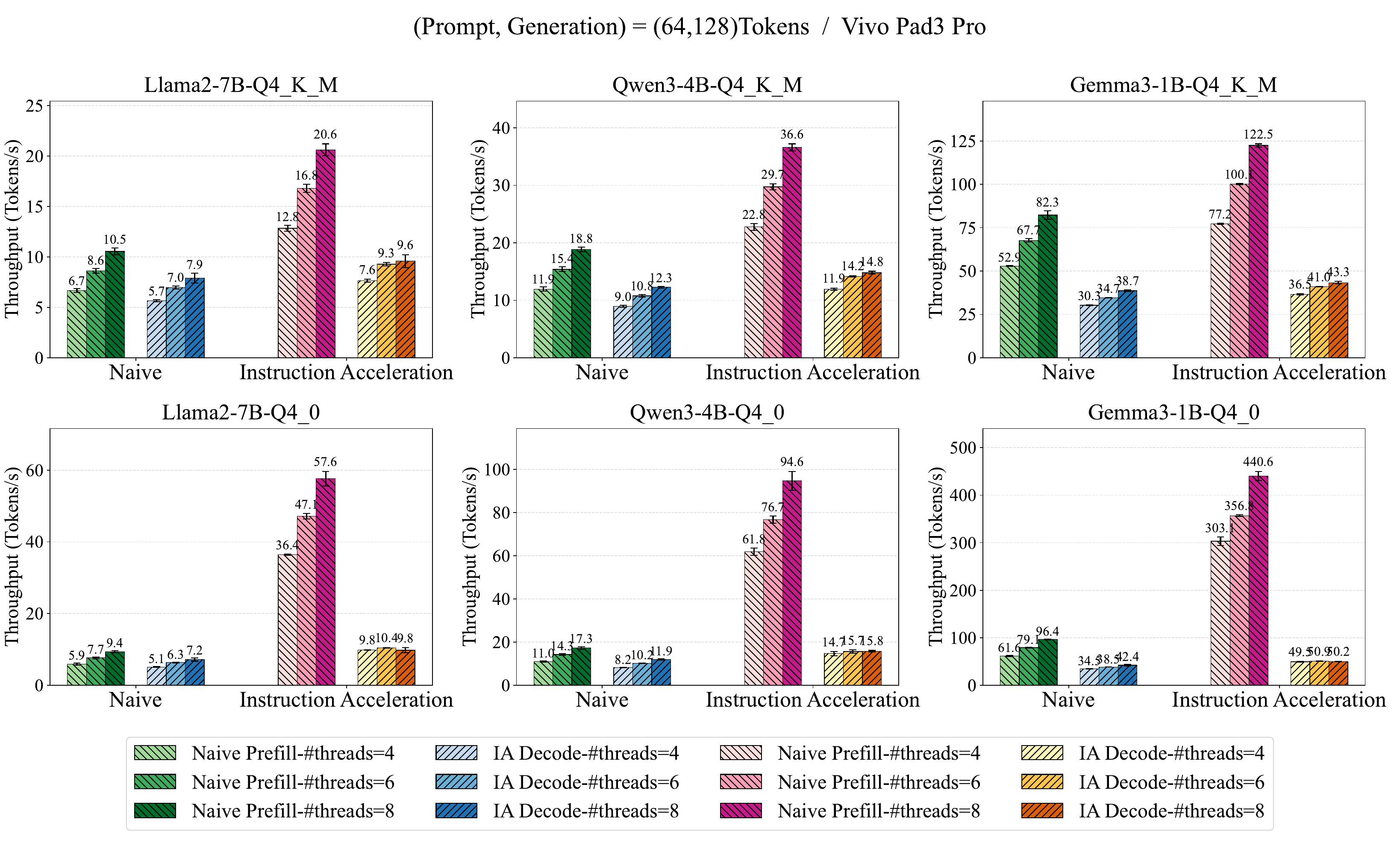}
      \vspace{-1.3em}
      \subcaption[]{\new{Vivo Pad3 Pro (Dimensity 9300)}}
      \label{fig:i8_throughput_vivo}
    \end{minipage}
    \caption{\new{Inference Performance with INT8 Matrix Multiplication (llama.cpp)}}
    \label{fig:i8_throughput}
\end{figure*}

We evaluate three core configurations: using only big cores, a combination of big and little cores, and all available cores. To enable parallel inference across multiple CPU cores, we adjust the number of running threads. Since inference threads are prioritized to run on big cores, the number of running threads implies which cores are active. For instance, on the Snapdragon 8 Gen 3 (which has six big cores), six threads correspond to all big cores, while on the Snapdragon 8+ Gen 1 (which has only four big cores), the same configuration includes all big cores and two little cores.

Figure~\ref{fig:multi-threads} illustrates how these core configurations impact inference speed across the four tested devices. During the prefill stage, prompt processing speed is primarily driven by the big cores. The contribution of the little cores to acceleration depends on the performance gap between the big and little cores. On devices with limited computational power, additional efficiency cores can help accelerate inference. For instance, utilizing all cores delivers the best performance on the Kirin 9000E.
However, for devices with more powerful big cores, adding little cores can actually degrade performance. For example, on the Snapdragon 8 Gen 3, incorporating two little cores results in a slowdown in inference speed. This highlights the importance of optimizing core configurations based on each device's specific CPU capabilities to maximize efficiency. Interestingly, the Dimensity 9300 benefits from using all cores because of its All-Big-Core architecture~\cite{dimensity9300}. %
\revision{On the Apple M1, we observe that even a single performance core achieves a prefill speed comparable to that of $8$ threads. Increasing the thread count does not improve CPU utilization; instead, it may introduce excessive context switching, where threads are frequently interrupted, leaving substantial idle intervals on CPUs and reducing CPU utilization.}

In the decoding phase, adding more little cores typically leads to a decline in performance. For instance, both the Kirin 9000E (Huawei Matepad12.6 Pro) and Snapdragon 8+ Gen 1 (Xiaomi Pad6 Pro) achieve their peak decoding speeds with four active big cores. This indicates that the maximum decoding speed on a device is largely dictated by the performance of the big cores. Unlike the prefill stage, decoding is more heavily constrained by memory bandwidth. This is evident in the fact that adding more big cores results in a smaller performance improvement compared to prefill, while incorporating little cores leads to a more noticeable performance drop.
\begin{observation}
To achieve optimal inference performance on multi-core CPUs, the decision to activate a core for inference should depend on its computational capacity relative to other cores. In general, the primary cores and performance cores contribute most to overall performance, while the efficiency cores have a negligible or even negative impact due to their limited computing resources. One potential method is distributing the workload to each core based on its compute capability and memory access overhead.   
\end{observation}

\begin{observation}
Since the prefill stage is compute-bound and the decoding stage is memory-bound, the core activation strategy should be adapted for each stage. Although scaling to more cores improves theoretical computation capacity, it risks introducing more synchronization and memory overhead. Take Kirin 9000E and Snapdragon 8+ Gen 1 as examples, the efficiency cores provide slight improvement during prefill, but degrade performance in decoding.
\end{observation}

\subsubsection{\new{Speedup with Special Machine Instructions}}
\label{subsec:instruction}
CPUs are generally more efficient at handling integer (INT) operations compared to floating-point (Float) computations. This performance advantage in INT operations stems from two key factors: first, CPUs typically offer higher instruction throughput for INT operations than for floating-point ones. Second, for matrix multiplication tasks, many CPUs support specialized INT8 instructions, which further accelerate these computations.

\new{To further explore the potential of Arm CPUs, we select three models of varying sizes and architectures: Gemma$3$-$1$B, Qwen$3$-$4$B, and Llama$2$-$7$B. We recompiled llama.cpp directly on top-tier devices with the armv$8.7$ flag to enable INT$8$ matrix multiplication instructions. This ensures that the machine code includes \textit{smmla}\cite{armsmmla} and other efficient Neon instructions. Figure~\ref{fig:i8_throughput} illustrates that enabling these special machine instructions significantly improves throughput, with the acceleration primarily observed in the prefill stage. While the decode stage is memory-bound, it still shows slight improvement.}
\new{The Dimensity $9300$ maintains optimal performance with all big cores, but when both using $6$ cores, the Snapdragon $8$ Gen$3$ outperforms the Dimensity $9300$. This performance advantage is likely due to the higher CPU frequency of the Snapdragon $8$ Gen$3$. Furthermore, when both SoCs are configured with optimal core setups, the Snapdragon $8$ Gen$3$ experiences a slightly higher performance boost from instruction acceleration compared to the Dimensity $9300$.}

\new{For the same quantization method, larger model sizes experience more significant acceleration. The Snapdragon $8$ Gen$3$ and the Dimensity $9300$ both show the highest speedup with Llama$2$-$7$B. Specifically, in the prefill stage, speedup is more than $6\times$ with Q$4$\_$0$ and about $2\times$ with Q$4$\_K\_M. For Gemma$3$-$1$B, the speedup is around $5\times$ with Q$4$\_$0$ and $1.5\times$ with Q$4$\_K\_M.} %

\new{Compared to Q$4$\_K\_M, Q$4$\_$0$ optimizes weight layout by rearranging weights in blocks~\cite{arm2024llmcpu}. This approach eliminates pseudo-scalar operations and fully leverages the parallel capabilities of specialized instructions. While the original weight layout processes a single column at a time, the optimized layout distributes weight columns across multiple lanes, enabling parallel processing and reducing memory access, thereby enhancing inference efficiency. For the Snapdragon 8 Gen3, the presence of small cores slows down throughput, especially with Q4\_0. The performance gap between 6 and 8 cores widens with Q4\_0, likely due to the mismatch between the specialized acceleration instructions (paired with prearranged weights) and the architectural constraints of the efficiency cores (\ie, small cores). Interestingly, when using Q4\_0 with instruction acceleration, the Snapdragon 8 Gen3’s performance with 6 big cores outperforms the Dimensity $9300$’s 8 big cores, indicating that the specific operator implementation plays a crucial role in performance.} %

\begin{observation}
Software developers must consider the instruction set of the target platform and optimize the code. It can involve rewriting kernel functions, adjusting the memory layout, and tailoring the computation operations, which contributes to improving performance using specialized hardware instructions. On the other hand, hardware developers should focus on designing more efficient instructions to enhance the parallelism of matrix computations and improve memory utilization.
\end{observation}

\subsection{GPU Utilization for LLM inference}
\label{subsubsec:gpu_utilization}
We note an intriguing exception regarding performance on Mali GPUs. The Mali-G$720$ \delete{in the Dimensity 9300 (Vivo Pad3 Pro)}, has better hardware specifications but poorer inference performance than Adreno $750$ (Xiaomi14 Pro) \delete{in Snapdragon SoC}. We measure the maximum throughput for FP16 operations on devices using clpeak\cite{clpeak}, \revision{which is a benchmarking tool intended for calculating GPU performance by executing different OpenCL kernels. The actual maximum memory bandwidth is also obtained using clpeak and presented with the theoretical maximum bandwidth in Table~\ref{tlb:gpu_specifications}. }
From Table~\ref{tlb:gpu_specifications}, we can see that the Mali-G720 MP12 exhibits $1.5$ times the throughput of the Adreno $750$ in float16 operations. \new{However, the results in Figure~\ref{fig:gpu_throughput} reveals a stark contradiction, as the prefill speeds of the four tested models are approximately $10$ times slower on the Mali GPU than on the Adreno $750$.}

\begin{figure}[!t]
\centering
	\subfloat[\revision{Utilization of GPU units}]{\includegraphics[width = 0.48\textwidth]{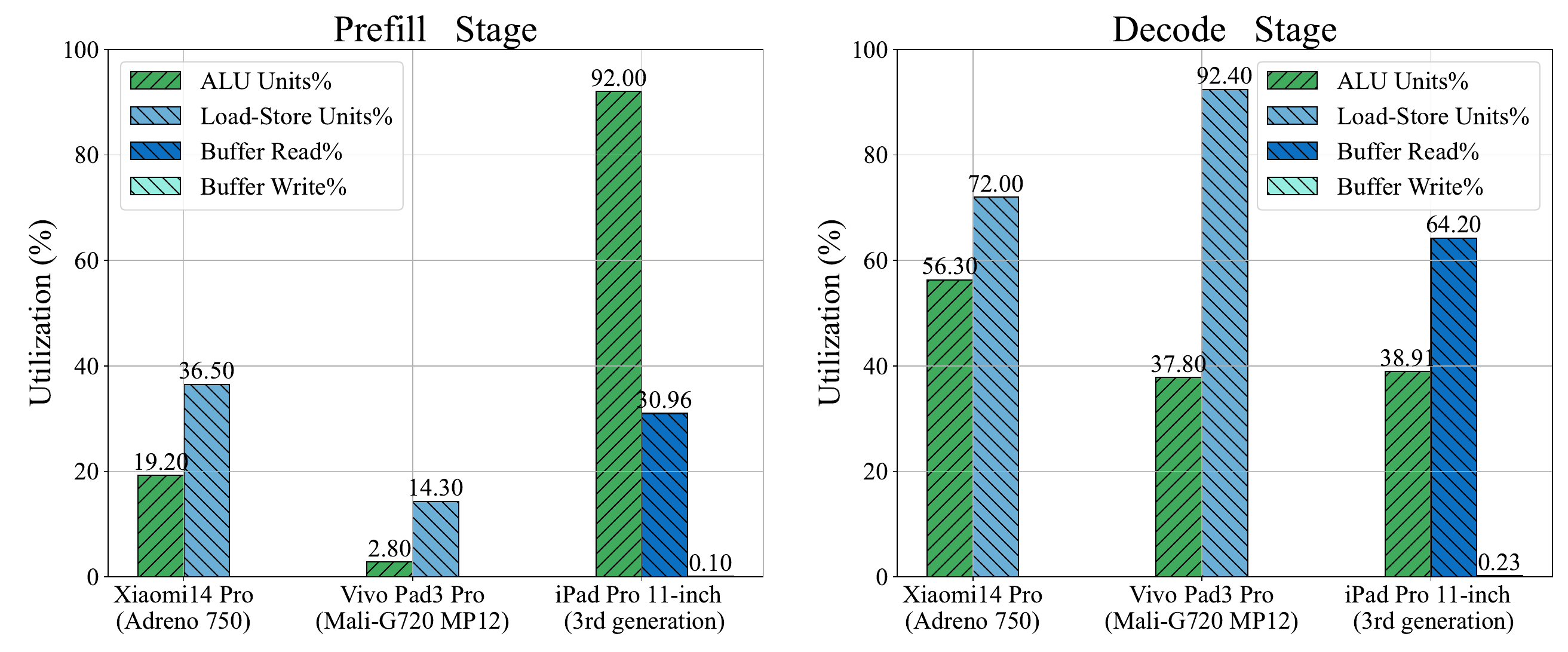}}
	\hfill
	\subfloat[\revision{Memory Bandwidth}]{\includegraphics[width = 0.48\textwidth]{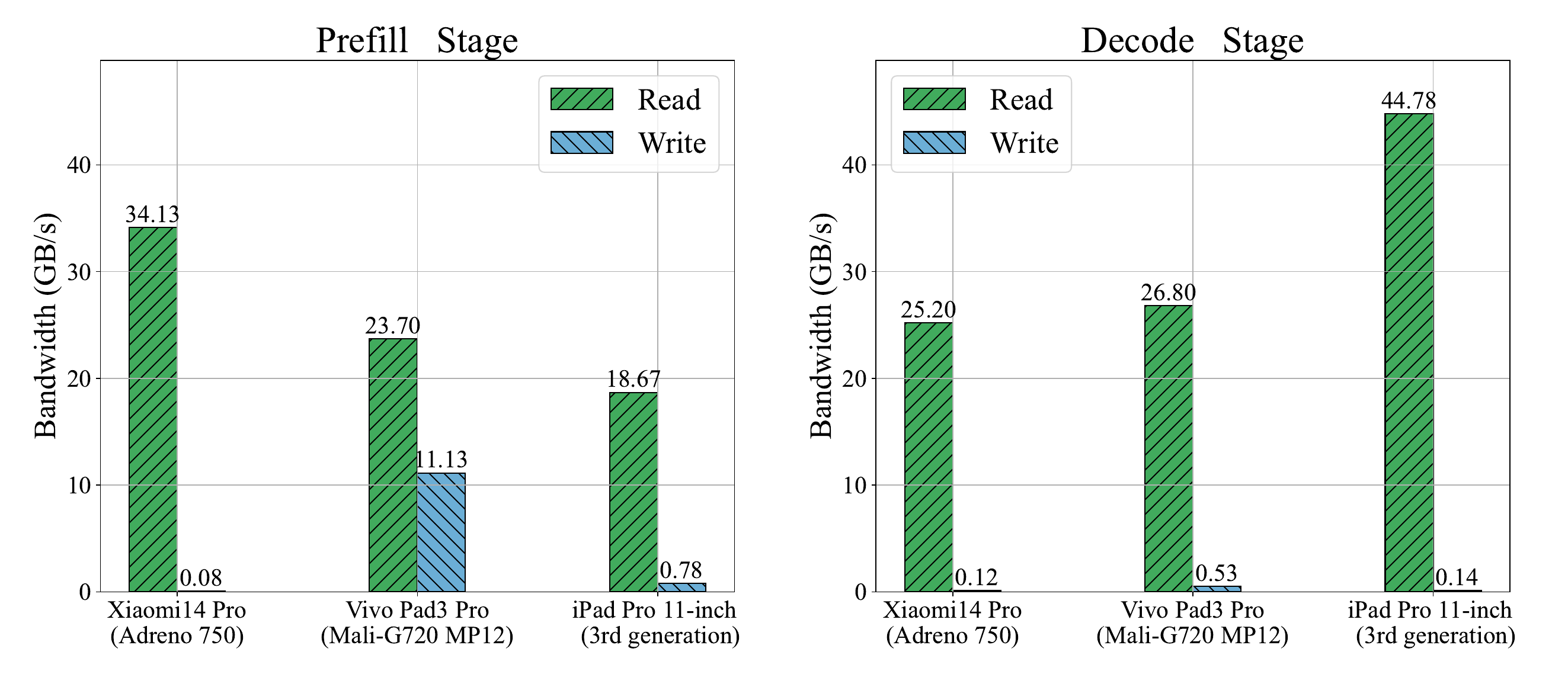}} 
\caption{GPU Utilization \new{(MLC LLM with Llama$2$-$7$B)}}
\label{fig_E2}
\end{figure}

\new{Using Llama2-7B as a representative case, we further analyzed a set of GPU performance metrics during inference to identify the root causes of this performance gap.} Figure \ref{fig_E2} presents the average utilization and memory bandwidth during inference on Mali-G$720$, Adreno $750$, and \new{Apple M$1$}. \revision{It should be clarified that, on Apple devices, Load-Store utilization is reported separately as buffer reads and buffer writes, whereas on non-Apple devices we refer to the aggregate Load-Store utilization. }%

\new{The possible reason for this discrepancy comes from the GPU utilization. Referring to Figure~\ref{fig:gpu_throughput}  and Figure~\ref{fig_E2}, the ALU utilization during the compute-bound prefill stage demonstrates a strong positive correlation with throughput. The Mali-G$720$’s arithmetic unit utilization averages less than $3$\%, implying that the matrix multiplication operators are almost not parallelized. The Adreno $750$ performs better, but only reaches around $20\%$. In contrast, the Apple M$1$ achieves up to $92$\%, providing approximately an $8\times$ speedup over the Adreno $750$.}
In the memory-bound decoding stage, the Mali-G$720$ shows a severe memory bottleneck for the nearly saturated Load-Store units utilization. Taken together, these findings highlight that generic operator implementations are mismatched with the Mali GPU, leaving compute resources underused in the prefill stage and memory resources overwhelmed in the decode stage.

\subsubsection{ \revision{Understand the underuse of non-Apple GPUs}}

\revision{
To further investigate the causes of GPU underutilization and explore optimization opportunities, we conduct additional experiments and analyze a broader set of metrics on the Adreno $750$ and Mali-G$720$. The results consistently highlight that the primary factors contributing to underutilization are the inefficient memory access pattern and the lack of operator optimizations tailored to specific GPU architectures.
}

\begin{table}[!tb]
  \centering
  \small
  \caption{\revision{GPU specifications and corresponding  achievable performance}}
  \resizebox{0.46\textwidth}{!}{\begin{tabular}{ccc}
    \toprule
    GPU Name & \revision{FP16} & Est./Theoretical Max  \\
    & (GFLOPS) &Memory Bandwidth (GB/s) \\
    \midrule
    Mali-G78 MP22 &  $2065$ & $26$/- \\
    \midrule 
    Mali-G720 Immortalis MP12 & $6456$ &  $48$/77 \\
    \midrule
    Adreno 730 & $3086$ & $39$/- \\
    \midrule
    Adreno 750 & $4314$ & $63$/77\\
    \bottomrule
  \end{tabular}}
  \label{tlb:gpu_specifications}
\end{table}

\begin{figure}[!tb]
  \centering
  \includegraphics[width=0.7\linewidth]{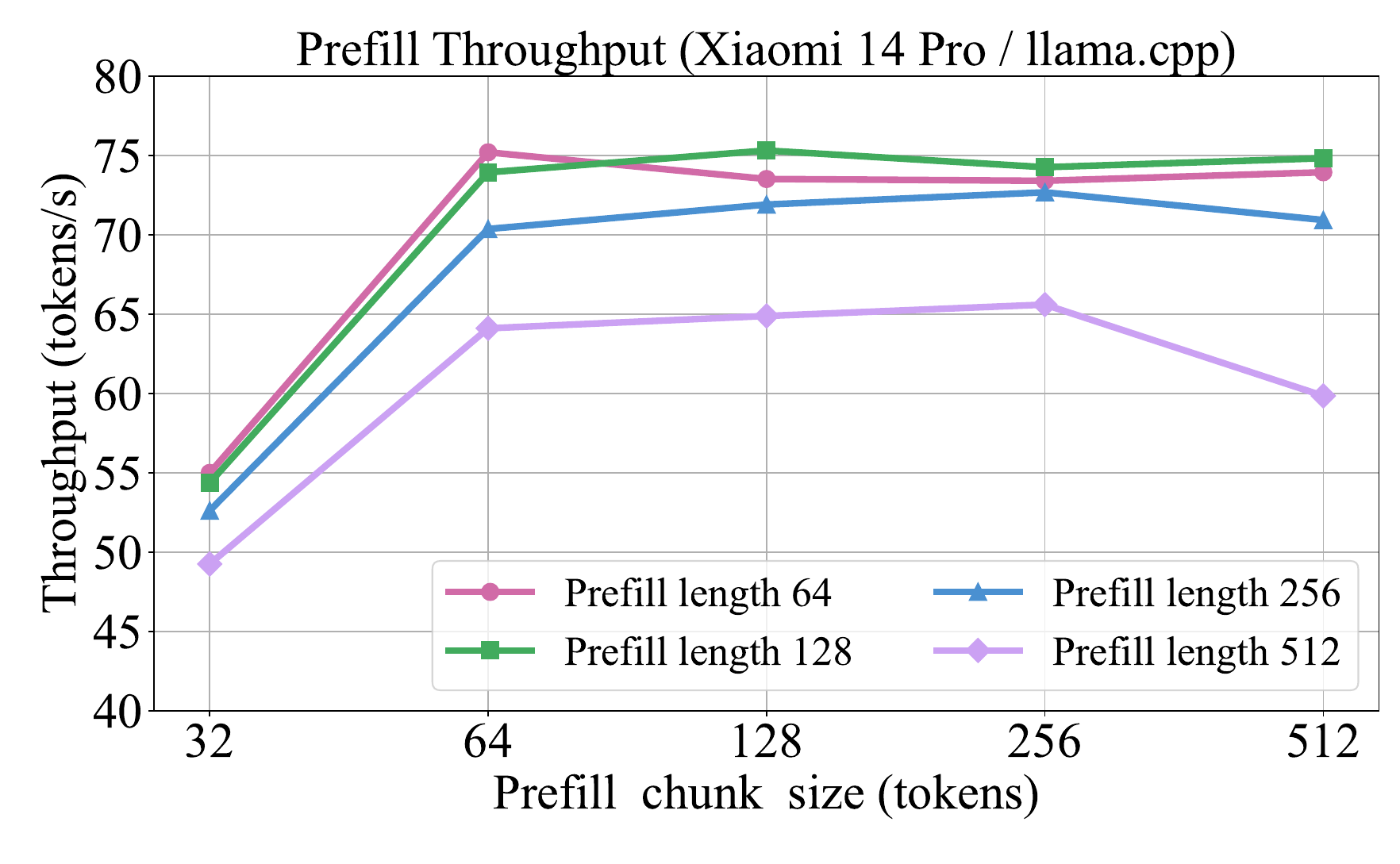}
  \caption{\revision{Prefill Throughput under various chunk lengths}}
  \label{fig:prefill_chunk}
\end{figure}

\new{During the prefill stage on the Adreno $750$, profiling reveals that ALU underutilization is mainly caused by memory access stalls. In a single layer of Llama$2$-$7$B, the final matrix multiplication kernel experiences up to $60$\% of its cycles stalled due to  memory access, with L$2$ cache miss rates exceeding $60$\%. This indicates that the bottleneck probably results from non-contiguous memory access, which causes prolonged L$2$ stalls and reduces ALU activity. For other kernels, approximately $40$\% of the stalls are attributed to texture memory access, indicating that the bottleneck arises from the volume of texture memory requests exceeding the available hardware bandwidth. This suggests that the inefficiencies may be caused by redundant data requests or suboptimal data sharing between threads, which prevent the kernel from effectively reusing cached data.}

On the Mali-G$720$, a similar problem arises: over $80\%$ of Load-Store units reads are partial-width, resulting in numerous short memory accesses that underutilize cache capacity~\cite{armpartial}. Moreover, the rate of all registers warp reaches nearly $100\%$, which halves the peak thread occupancy of the shader core and makes shader performance particularly sensitive to cache misses and memory stalls\cite{armwarp}. Taken together, the findings demonstrate that, although the prefill stage is inherently compute-intensive, suboptimal memory access patterns constitute the primary performance bottleneck in existing implementations.

\revision{
 As for the decode stage, the memory-bound nature of operators (i.e., matrix–vector multiplication) leads to frequent memory accesses for matrix operands. On the Adreno $750$, this is reflected by an L$2$ cache miss rate approaching $100\%$. The issue is even more pronounced on the Mali-G$720$, where the average load-store unit utilization exceeds $90\%$, with nearly all read operations being partial-width. Moreover, more than $60\%$ of external memory accesses exhibit high latency (over $256$ cycles), suggesting that the memory system is struggling to keep up with the requested data volume~\cite{armlatency}. These findings suggest that optimizing data access patterns to reduce partial-width reads and alleviating bandwidth pressure are critical to mitigating memory bottlenecks in the decode stage.}

\subsubsection{ \revision{GPU Optimization Practices}}

\revision{ Qualcomm’s optimized version of MLC LLM demonstrates that careful data layout design can substantially enhance GPU efficiency~\cite{MLCv2}. By reorganizing the input matrix into $32\times16$ tiles with contiguous memory addresses, the implementation achieves improved memory coalescing. Profiling further shows that cache misses in L$1$ and L$2$ occur only at limited points (e.g., within the first $2$ ms of a $10$ ms execution). As a result, the prefill speed reaches over $50$ tokens/s, representing a six-fold improvement over the general version on the same device. This case shows that optimizing memory access patterns alone can yield substantial prefill performance gains.}

\revision{
Additionally, the absence of customized operator implementations further limits GPU utilization. In MLC-LLM, all operators are cross-compiled uniformly for mobile devices, resulting in identical parameter configurations for both Adreno and Mali GPUs. However, since these GPUs differ in core counts, cache capacities, and memory bandwidth, parameters of OpenCL kernels such as global work size should be tuned and optimized individually through empirical testing. 
}

\revision{To verify the impact of kernel parameter, we evaluate different prefill chunk sizes using llama.cpp on the Xiaomi $14$ Pro. Specifically, the OpenCL kernels in llama.cpp use a fixed local work size~\cite{armworksize}, while the global block size is determined by the prefill chunk size. Large blocks typically cause memory contention, whereas small blocks underutilize compute units. Figure~\ref{fig:prefill_chunk} illustrates the impact of chunk size on throughput for different prompt lengths. On the Xiaomi 14 Pro, prompts longer than 256 tokens achieve peak performance at a chunk size of 256, while shorter prompts perform best when the chunk size matches the prompt length. These optimal values, however, may vary across devices, suggesting that pre-testing or profiling is necessary to determine device-specific configurations.}

\begin{observation}
General-purpose LLM implementations on mobile GPUs do not fully harness the parallelism capabilities of these devices. 
The results highlight a significant performance gap between different GPU architectures (e.g., Mali vs. Adreno)—for instance, Adreno GPU may achieve \new{$10\times$} throughput than Mali GPU (which has higher computation capacity) for the same workload. %
To unlock the full potential of mobile GPUs, customized implementations are essential.
\end{observation}

\revision{
\begin{observation}
The primary performance bottleneck on GPUs arises from suboptimal data layouts and memory access patterns. Even in the compute-bound prefill stage, frequent external memory accesses and \new{overloaded texture memory requests} cause significant ALU idle time. Optimizations such as restructuring data layouts or applying memory coalescing can improve cache utilization, while rebalancing thread workloads and minimizing register usage can further enhance computational efficiency.
 \end{observation}}

\begin{table}[!t]
  \centering
  \footnotesize 
  \caption{\new{Power Consumption ([Prompt, Gen] = [$64$, $128$])}}
  \label{tlb:energy_consumption_compact}
  
  \setlength{\tabcolsep}{4pt} 
  
  \renewcommand{\arraystretch}{1.1}
  
  \begin{tabular}{@{} l c cc cc @{}}
    \toprule
    \multirow{2}{*}{\makecell[l]{\textbf{Device} \\ \textbf{(CPU/GPU)}}} & \multirow{2}{*}{\textbf{\#Threads}} & \multicolumn{2}{c}{\textbf{Prefill}} & \multicolumn{2}{c}{\textbf{Decode}} \\
    \cmidrule(lr){3-4} \cmidrule(lr){5-6}
    & & Tput\textsuperscript{1} & Power\textsuperscript{2} & Tput\textsuperscript{1} & Power\textsuperscript{2} \\
    \midrule
    \textbf{Vivo Pad3 Pro} & $4$ & $6.11$ & $0.050$ & $5.01$ & $0.053$ \\
    (CPU w/o IA\textsuperscript{3})           & $6$ & $8.08$ & $0.039$ & $6.40$ & $0.047$ \\
                           & $8$ & $9.80$ & $0.031$ & $7.09$ & $0.042$ \\
    \midrule
    \textbf{Xiaomi Pad6 Pro}& $4$ & $5.55$ & $0.106$ & $4.11$ & $0.137$ \\
    (CPU w/o IA)           & $6$ & $5.89$ & $0.106$ & $4.12$ & $0.143$ \\
                           & $8$ & $6.08$ & $0.106$ & $4.11$ & $0.143$ \\
    \cmidrule{2-6}
    Xiaomi Pad (CPU w/ IA)     & $6$ & $12.87$ & $0.038$ & $4.51$ & $0.117$ \\
    \cmidrule{2-6}
    Xiaomi Pad (GPU)           & $6$ & $46.03$ & $0.013$ & $7.33$ & $0.058$ \\
    \midrule
    \textbf{Huawei Matepad12.6 Pro}& $4$ & $4.28$ & $0.100$ & $3.56$ & $0.118$ \\
    (CPU w/o IA)           & $6$ & $4.79$ & $0.137$ & $3.43$ & $0.111$ \\
                           & $8$ & $4.99$ & $0.144$ & $3.60$ & $0.120$ \\
    \bottomrule
    \multicolumn{6}{@{}l@{}}{\scriptsize $1$: Throughput (tokens/s); $2$: Power (mAh/token); $3$: Instruction Acceleration}
  \end{tabular}
  \vspace{-10pt}
\end{table}

\subsection{\new{Battery Consumption}}
\new{While inference speed remains the primary metric for ensuring high-quality user experience, power consumption is equally critical, as it directly determines the feasibility of sustaining on-device LLM inference. To comprehensively evaluate the trade-off between throughput and energy efficiency, we evaluated three representative mobile devices across distinct performance tiers with llama.cpp and Llama$2$-$7$B. The experimental results represent the mean of two independent trials, each consisting of 20 consecutive inference iterations with a 64-token prefill and 128-token generation. For precise energy measurement, we directly sample current and voltage from system files on the rooted Xiaomi device. For non-rooted devices, energy metrics were obtained from application-specific power consumption data provided by the operating system's monitoring tools. The results are shown in Table~\ref{tlb:energy_consumption_compact}.}

\new{\textbf{\textit{Core Configuration and Efficiency}}: Adjusting the number of threads influences energy-to-throughput efficiencies across different SoC architectures. For top-tier devices, such as the Vivo Pad$3$ Pro which utilizes an all-big-core architecture, increasing the thread count improves efficiency. As more cores are engaged, inference speed increases, and energy efficiency improves. For example, when using all cores, prefill energy consumption drops to its lowest point at 0.031 mAh per token. In contrast, high-tier devices featuring a combination of big and little cores, such as the Xiaomi Pad$6$ Pro, experience diminishing returns—adding more cores provides only a modest speed improvement for prefill, but power consumption remains unchanged. For devices with a weaker big-little core configuration, such as the Huawei tablet, performance gains from multi-threading come at the expense of higher power consumption, particularly during the prefill stage, exhibiting significant energy overhead.} 

\new{\textbf{\textit{Prefill vs. Decode Energy Profile}}: Energy consumption per token shows distinct patterns depending on the device’s hardware capability. Typically, the decode stage is more energy-intensive due to its sequential, auto-regressive nature, which doesn't benefit from batch parallelization. However, on less powerful devices, this pattern reverses. The Huawei tablet is the only device tested where prefill energy exceeds decode energy. As the thread count increases from 4 to 8, prefill energy rises from 0.1mAh/token to 0.14mAh/token, while decode energy remains nearly unchanged at approximately 0.12mAh/token.}

\new{\textbf{\textit{Hardware Acceleration and Energy Efficiency}:} Compared to standard CPU execution, specialized hardware offers transformative efficiency. On the Xiaomi Pad 6 Pro (using 6 threads), Instruction Acceleration (IA) (see Section~\ref{subsec:instruction}) speeds up prefill by $2.19\times$ while reducing energy consumption by $63.82$\%. The GPU shows even more dramatic improvements, accelerating prefill by $7.81\times$ and cutting energy consumption by $87.79\%$. Even in the less computational-intensive decoding phase, the GPU achieves a $1.78\times$ speedup while reducing energy consumption to 0.058mAh/token, a $59.75\%$ decrease. These results underscore the significant energy efficiency advantage of GPU-based inference compared to CPU execution.}

\begin{figure*}[!htb] %
    \centering
    \begin{minipage}[b]{0.55\textwidth}
       \centering
      \includegraphics[width=\linewidth]{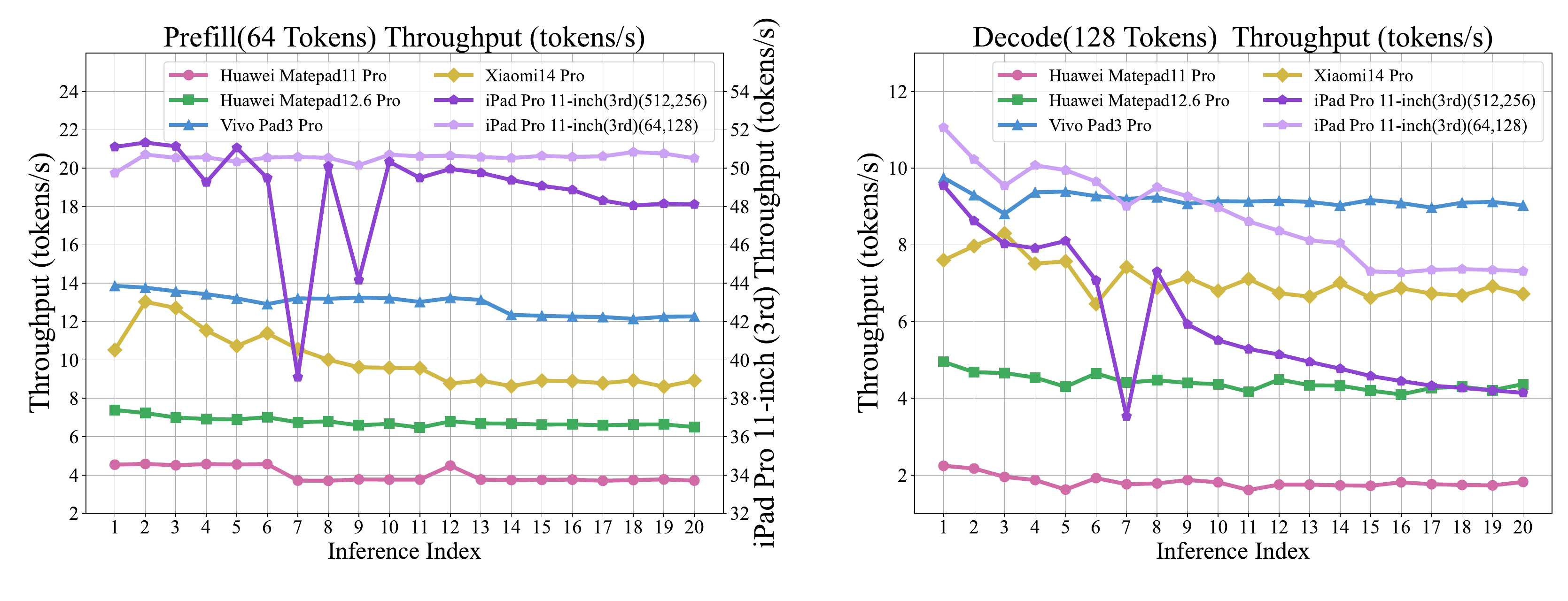}
      \caption{\revision{Throughput under continuous inference (CPU)}}
      \label{fig:DVFS}
    \end{minipage}
    \begin{minipage}[b]{0.25\textwidth}
        \setlength{\abovecaptionskip}{1pt}
      \centering
      \includegraphics[width=\linewidth]{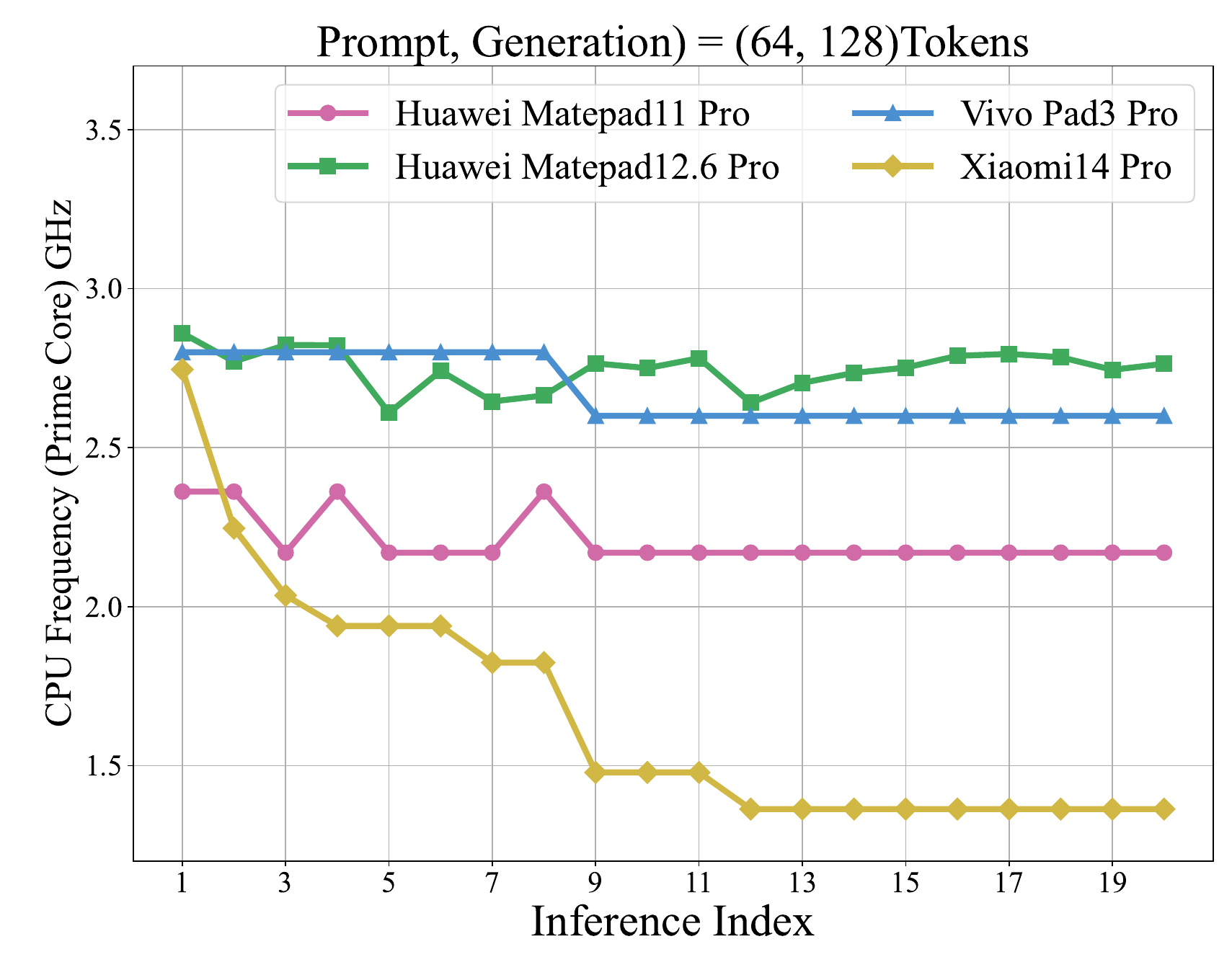}
      \vspace{0.1mm}
      \caption{\revision{Prime Core Frequency}}
      \label{fig:frequency}
    \end{minipage}
\end{figure*}

\subsection{Impact of DVFS and Thermal Throttling}

\subsubsection{Impact of DVFS \revision{on CPU}}
\label{subsec:dvfs}

In Section~\ref{subsubsec:cpu}, we observed performance degradation as the length of the prompt and generation increased. This issue relates to the system's ability to handle long contexts and maintain consistent inference performance across multiple rounds of dialogue. %
We analyze real-time variations in CPU frequency and utilization during 20 continuous inference tests on four devices with warm up, covering different operating systems (Android, HyperOS, HarmonyOS, \revision{iPadOS}) and vendors (MediaTek, Qualcomm, Hisilicon, \revision{Apple}). \revision{Given the significant performance advantage of the Apple device, we performed additional testing using longer prompts and outputs.}

Figure~\ref{fig:DVFS} illustrates a general decline in performance as the rounds of inference increase. %
\revision{Although the inference throughput for both prompt lengths is similar, %
the iPad exhibits a more pronounced performance drop in long-prompt tests compared to short-prompt tests. Moreover, in short-prompt tests, unlike the stable prefill speed, the decode speed on the iPad decreases during the early stages and eventually falls from the best to a level comparable with the second-highest among the non-Apple devices. Additionally, on Xiaomi 14 Pro, throughput for both prefill and decode initially rises, then decreases, and finally stabilizes at a relatively steady level, accompanied by minor oscillations around the equilibrium.} %
In contrast, Dimensity $9300$ (Vivo Pad3 Pro) and Kirin $9000$E (Huawei Matepad12.6 Pro) exhibit only minor latency fluctuations, with a maximum decrease of approximately $10\%$. This trend aligns with the data presented in Figure~\ref{fig:llamp_throughput}, which shows that Xiaomi $14$ Pro experiences a more significant performance drop as prompt length increases. %

Figure~\ref{fig:frequency} shows the frequency variation on the prime CPU core at different rounds of inference. \revision{CPU frequency is inaccessible on the iPad, resulting in a lack of corresponding measurement data.} \rev{The results reveal that Snapdragon SoCs (\eg, Huawei Matepad11 Pro, Xiaomi14 Pro) exhibit a more aggressive DVFS policy}. \delete{During subsequent stages of the inference, both prime and performance cores on Snapdragon 8 Gen 3 experience a rapid reduction in frequency.} By the $9$th inference round, the frequency is nearly halved compared to its initial value. To investigate the factors influencing DVFS policy, we further compare two Huawei tablets equipped with identical CPU cores but different SoCs—one powered by Snapdragon (\delete{Snapdragon 870 on} Huawei Matepad11 Pro) and the other by Kirin (\delete{Kirin 9000E on} Huawei Matepad12.6 Pro). The results indicate that the vendor plays a significant role in shaping the DVFS policy. For instance, %
\revision{the Snapdragon 870 typically keeps the prime core frequency lower than the Kirin 9000, although they have the same maximum frequency.}

\begin{observation}
CPU frequency critically affects inference performance. The default DVFS strategy tends to be conservative, taking factors such as thermal throttling and battery life into consideration. While this strategy is suited for successive multi-round inferences, the more common case on mobile devices involves intermittent single-round inference (with larger than $30$s interval). Therefore, developers may consider increasing the CPU frequency—or even overclocking—for initial inference to reduce response latency.
\end{observation}

\subsubsection{\revision{Impact of CPU Thermal Throttling}}
\label{subsec:thermal}

\revision{
As discussed in Section~\ref{subsec:dvfs}, only a few consecutive inference runs are sufficient to trigger CPU frequency reduction due to thermal throttling. To further examine the impact of heat accumulation over time, we extend the experiments on Xiaomi $14$ and analyze ($1$) temperature and performance variation during sustained inference and ($2$) the impact of initial device temperature on performance.}

\begin{figure*}[htb] %
    \centering
    \begin{minipage}[b]{0.55\textwidth}
       \centering
      \includegraphics[width=\linewidth]{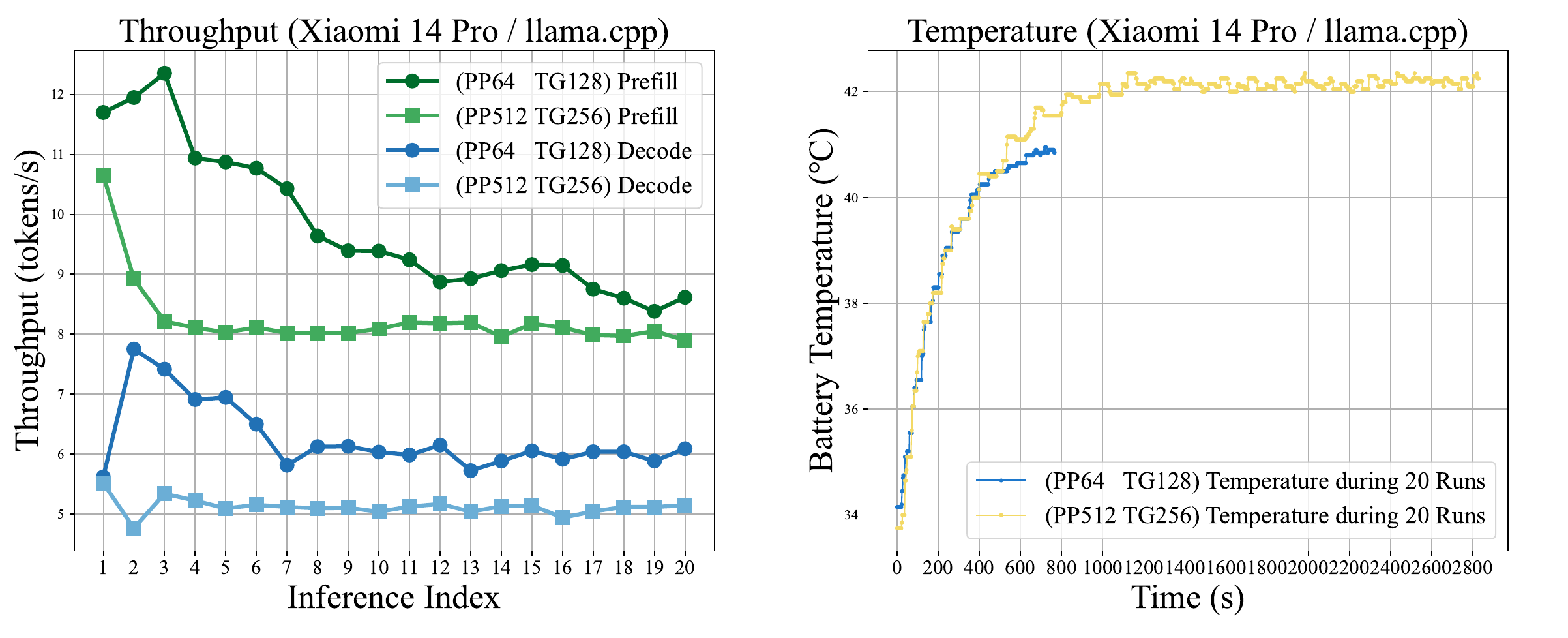}
      \caption{\revision{Throughput under continuous inference}}
      \label{fig:temp}
    \end{minipage}
    \begin{minipage}[b]{0.25\textwidth}
        \setlength{\abovecaptionskip}{1pt}
      \centering
      \includegraphics[width=\linewidth]{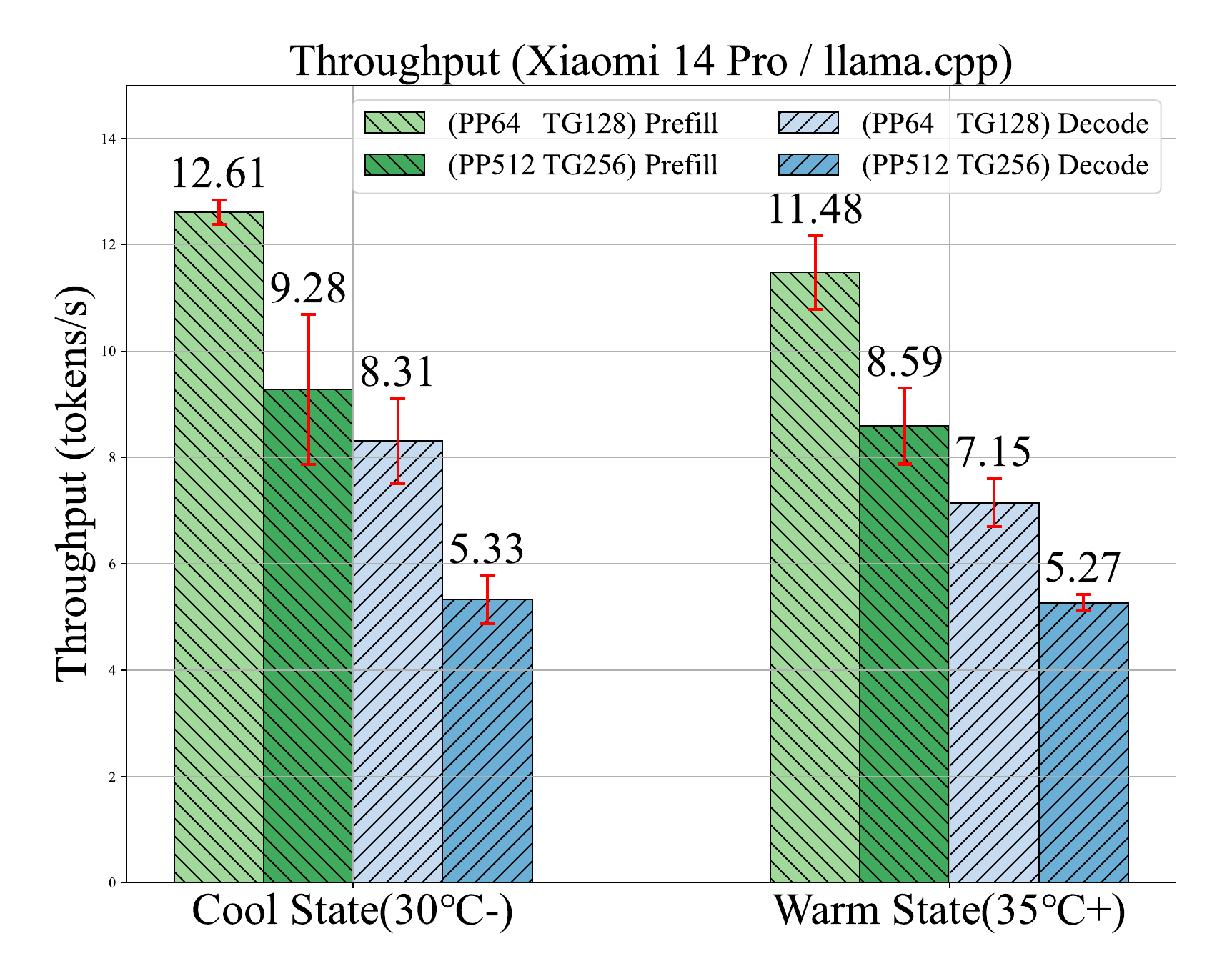}
      \caption{\revision{Performance Sensitivity to Starting Temperature}}
      \label{fig:temp_sens}
    \end{minipage}
\end{figure*}

\begin{figure*}[!t]
  \centering
  \includegraphics[width=0.8\linewidth]{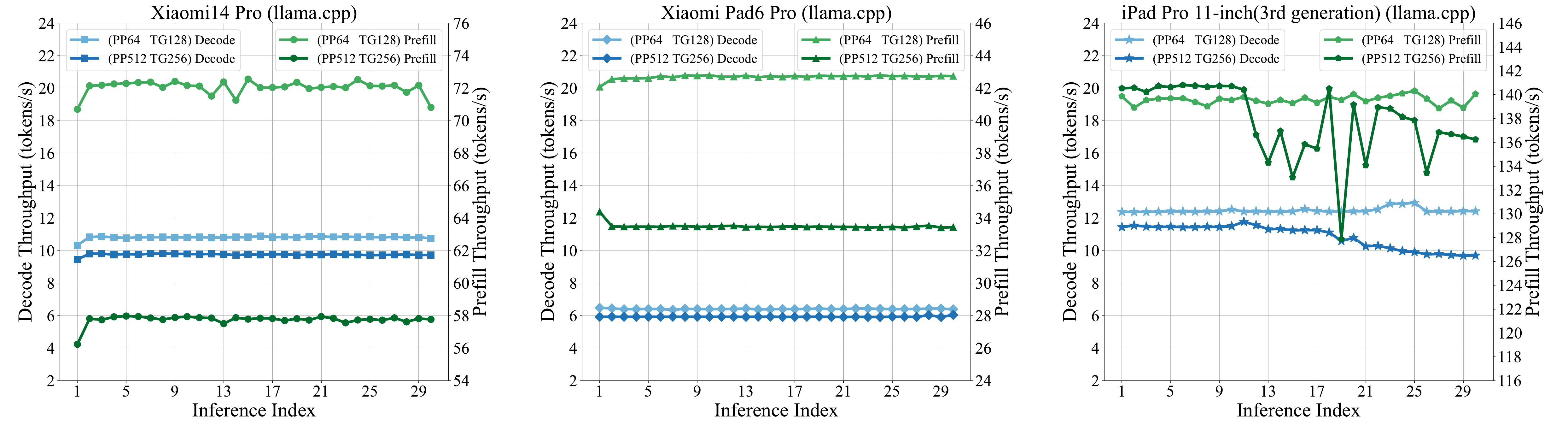}
  \caption{\revision{Throughput under continuous inference (GPU)}}
  \label{fig:GPU_DVFS}
\end{figure*}

\revision{\textit{Impact of Heat Accumulation on Sustained Inference:}}
\revision{
Figure~\ref{fig:temp} illustrates performance variations of different workloads over $20$ consecutive inference runs and the corresponding thermal buildup. Both curves exhibit an initial rapid temperature rise followed by stabilization. During the first $500$ seconds, the trends are nearly identical, starting from about $34^\circ\text{C}$ and climbing above $40^\circ\text{C}$. \new{In the stable stage, the rate of temperature rise slows, but the long-prompt tests still experience a faster rate of temperature rise compared to the short-prompt tests, }leading to a higher equilibrium near $42^\circ\text{C}$. After about $1000$ seconds, the temperature of long-prompt tests fully stabilizes, with minor oscillations around equilibrium likely reflecting stricter thermal regulation.
}

\revision{
The performance variation closely follows the temperature trend. During the initial heating stage, throughput fluctuates noticeably. For long-prompt tests, since each inference run takes longer, performance becomes consistent after only $2$–$3$ runs. In contrast, short-prompt tests stabilize much later, requiring nearly $10$ runs. Across both prefill and decode stages, the performance gap between the two tests is initially narrow, then widens as the system heats up, and finally narrows again and stabilizes at a steady difference of about $1$ token/s.
}

\textit{\revision{Performance Sensitivity to Starting Temperature:}}
\revision{ To evaluate performance sensitivity to starting temperature, experiments are conducted under two conditions: a cool-down state (below $30^\circ$C) and a warm-up state (above $35^\circ$C). These conditions are achieved by first running additional tasks to heat the device and then cooling it for varying durations until the target temperature is reached. All other experimental configurations are largely consistent with Section~\ref{method}. }

\revision{Figure~\ref{fig:temp_sens} shows that overall performance decreases at higher temperatures, with smaller workloads being more sensitive. Although temperature initially rises at a similar rate across different prompt lengths, long-prompt tests require more time per run, causing the CPU frequency to drop rapidly within the first one or two rounds of interference. Consequently, the effect of the initial temperature for long-prompt tests is largely limited to these early rounds, whereas for short-prompt tests, all five consecutive runs occur during the temperature rise stage, allowing lower starting temperatures to delay CPU throttling and sustain higher performance throughout.}

\subsubsection{\revision{Impact of DVFS on GPU}}
\label{subsec:dvfsgpu}

\revision{
We further evaluate the impact of DVFS on GPUs, using llama.cpp since continuous inference with MLC LLM often crashes. Due to framework limitation, only Xiaomi and Apple devices are tested and the results are illustrated in Figure~\ref{fig:GPU_DVFS}. Leveraging faster GPU inference with llama.cpp, experiments are extended to $30$ runs with larger workloads ($512$-token prompts, $256$-token outputs). GPU frequency data are available only on the rooted Xiaomi Pad $6$ Pro, while iPad reports thermal states via Xcode.}

\textit{\revision{Adreno GPU under Android:}} 
\revision{The results show that inference speed on both the Xiaomi $14$ Pro and Xiaomi Pad $6$ Pro remains relatively stable throughout the tests. GPU frequency on Xiaomi Pad $6$ Pro is generally fixed at $815$ MHz during inference. However, in the first run of some tests, it briefly spikes to $862$ MHz before quickly returning to $815$ MHz, where it remains stable for subsequent runs. This initial fluctuation is likely caused by the DVFS policy.}
    
\textit{\revision{Apple GPU under iOS:}} 
\revision{Unlike the stable performance observed on Adreno GPUs under Android, the iPad exhibits noticeable fluctuations throughout the inference process. For short-prompt tests, the performance degradation is minor and appears near the end of the run. However, for long-prompt tests, the degradation occurs earlier and becomes more pronounced. These results are consistent with the thermal state reported: during most of the 30 inference runs, the device remains in the ``nominal" state (within normal thermal limits), but towards the end of long-prompt inferences, the state may shift to ``serious", indicating elevated heat conditions.}

\revision{
\begin{observation}
Compared with the CPU, whose frequency fluctuates frequently, the GPU maintains a relatively stable frequency during inference, yielding more consistent performance. This stability likely benefits from the GPU’s higher computational efficiency and lower power consumption. These observations suggest that offloading LLM inference to GPUs is a practical and effective strategy for preserving stable real-time responsiveness on mobile devices.
\end{observation}
}

\revision{
\begin{observation}
Mobile GPUs may not fully utilize their theoretical computational capacity. For example, the Xiaomi Pad $6$ Pro GPU supports $13$ frequency levels ranging from $220$ MHz to $900$ MHz, yet its default operating frequency is $815$ MHz, the third-highest level.
\end{observation}
}

\revision{
\begin{observation}
The end-to-end inference speed suggests that Apple’s DVFS strategy for the GPU is more conservative than that of Adreno GPUs on Android. Although the Apple GPU offers higher computational capability, it is more prone to performance degradation under sustained workloads due to stricter frequency-scaling policies.
\end{observation}
}

\subsection{Model Preparation Latency}

\begin{figure}[!t]
  \centering
  \includegraphics[width=\linewidth]{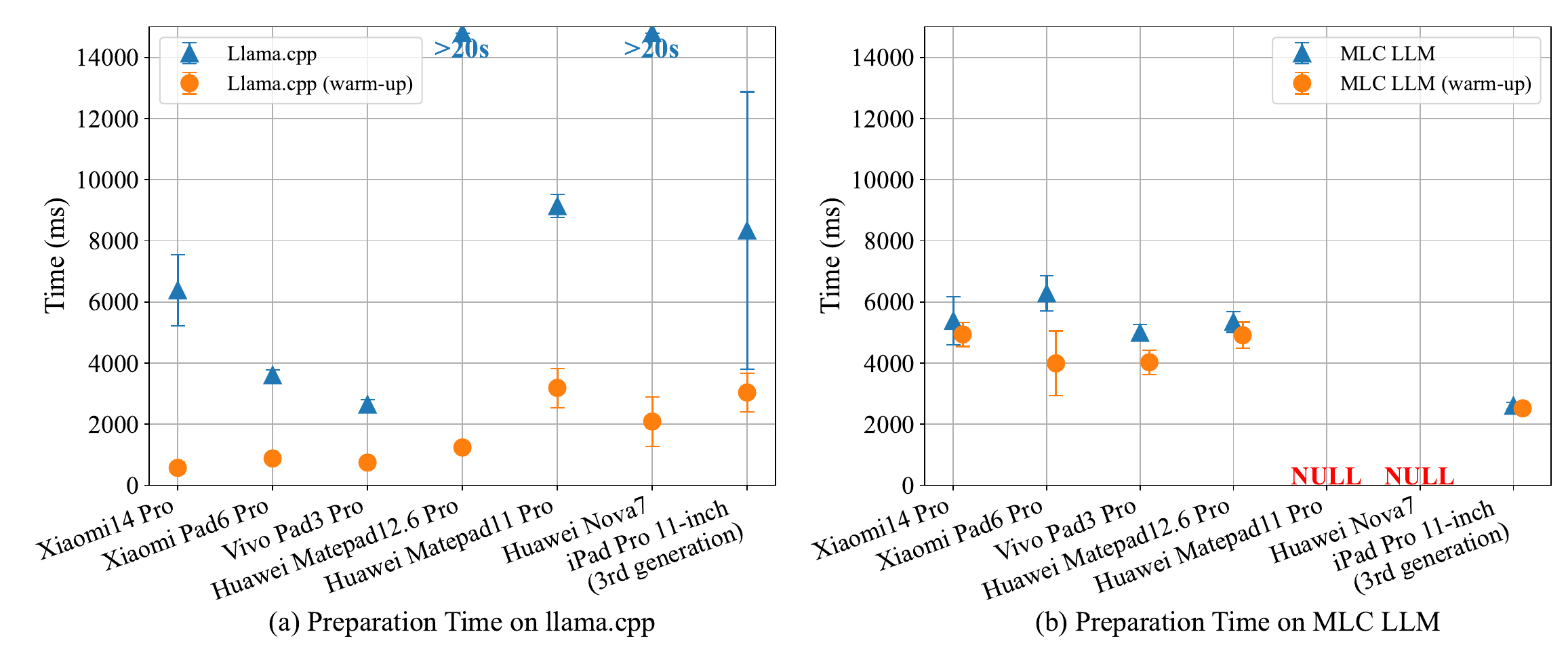}
  \caption{\revision{Time for model preparation}}
  \label{fig:preparation}
\end{figure}

In Section~\ref{subsec:dvfs}, we have studied the inference performance with warm-up. %
However, in real-world applications, users may trigger the model sporadically, leading to scenarios where the model must perform a cold start without a warm-up, known as the cold start. Thus, the time required to generate the first token becomes critical to user experience.  The process of generating the first token involves several stages: loading model weights from external storage into memory, preparing memory for variables, and executing prefill and decoding. In this section, we focus on the preparation procedure before prefill and evaluate the latency across various devices. We perform tests on \revision{seven} devices and measure the time spent on preparation with and without warm-up. If the device is tested without warm-up, we reboot it to refresh RAM and cache. Results are present in Figure~\ref{fig:preparation}.

For the llama.cpp running on CPUs, we compare the time before prefill of the first token in both cold-start and warm-up scenarios. \revision{The Xiaomi $14$ Pro performs the fastest after warm-up, while its preparation time for cold start is nearly ten times slower than for warm-up. The Vivo Pad $3$ Pro delivers the best overall performance for both warm-up and cold start. Additionally, the Apple device exhibits significant fluctuations during cold start, and its warm-up latency is noticeably slower than that of other top-tier devices.}

We recognize that the warm-up process significantly reduces preparation time for llama.cpp, primarily due to the use of \textit{mmap}, which allows the inference process to directly access model weight files from external memory. The \textit{mmap} function accelerates loading by eliminating the need to copy data between user and kernel space. Additionally, the mapping remains valid for other processes accessing the same file, meaning that after a warm-up, the model weights are already loaded into memory. This allows subsequent inference processes to use the preloaded weights, thereby reducing latency. %

From Figure~\ref{fig:preparation}, we observe that all Huawei devices exhibit unusually long preparation times during cold starts. On Huawei devices with Kirin SoCs, the delays are even more pronounced, with the worst loading times exceeding \revision{$20$} seconds. Huawei Matepad 11 Pro, despite using a Snapdragon SoC, also experiences extended delays. We identify the underlying causes for these exceptions are related to the \textit{mmap} function in Harmony OS. Through timestamp analysis, we identify that the primary cause is the execution of the \textit{mmap} function. On the Huawei Matepad 12.6 Pro, \textit{mmap} execution takes up to \revision{$20$} seconds, whereas after warm-up, this time is drastically reduced to \revision{$1.2$} seconds. This suggests that the delay arises from the memory mapping process rather than loading from UFS.

In contrast, for MLC LLM running on GPUs, no significant improvement is observed with warm-up, with loading times remaining approximately \revision{nine} times longer on Adreno GPUs and \revision{five} times longer on Mali GPUs compared to llama.cpp (with warm-up). \revision{An exception is that the Apple GPU performs slightly better than the CPU when using llama.cpp (with warm-up).}
\revision{This is probably due to the difference between the system memory architectures. For non-Apple devices, } \delete{although the CPU and GPU on mobile devices share the same physical DRAM, they operate in separate memory spaces and cannot directly access each other's data\cite{zhang2022comprehensive}. As a result, }model weights must be copied from the CPU memory space to the GPU memory space, introducing additional latency. Specifically, MLC LLM uses OpenCL kernels to transfer data between the CPU and GPU via buffers, which results in the delay. \revision{In contrast, a major advantage of Apple devices lies in their unified memory architecture~\cite{appleM1}, which enables shared access between CPU and GPU and thereby achieves much lower transfer latency.}

\section{Opportunities for Improving Mobile LLM Performance}
\label{sec:insights}
Building on the results present in the previous sections, this section highlights several key aspects that developers can leverage to significantly enhance mobile LLM performance.

\vspace{0.05in} \noindent \textbf{Hardware Instructions}
A key aspect of SoC evolution is the continuous enhancement of instruction sets. New machine instructions often increase operand width, resulting in higher throughput. To fully harness the potential of hardware, developers must consider the characteristics of model operators and take advantage of hardware-accelerated instructions.  This may also involve adjusting computational precision, such as quantizing activations to $8$-bit to utilize  instructions like \textit{smmla}. In Section~\ref{subsec:instruction}, we have discussed the improvement brought by SVE instructions. For next-generation CPUs based on the Armv9 architecture, advanced vector instructions like SVE2\cite{armsve2}, which enables longer computation vectors, are supported. Furthermore, Arm's SME instructions\cite{armsme}, specifically designed for matrix multiplication, present additional  optimization opportunities. Staying current with these advancements is crucial for developers seeking to maximize inference speed on mobile platforms.

\vspace{0.05in} \noindent \revision{\textbf{Dynamic Resource Allocation}
Parallel execution of large language models (LLMs) alongside other tasks remains a significant challenge. Optimal resource allocation is inherently dynamic, influenced by real-time workloads and system thermal conditions. A practical approach is to pre-define multiple operating modes that can adapt to varying workloads, such as adjusting the number of threads or offloading computations to idle processors.
}

\vspace{0.05in} \noindent \textbf{Faster First Generation}
The time required for generating the first token includes model preparation, prefilling and one round of decoding. Since the majority of model preparation involves I/O operations (loading model weights into memory), storing some of the weights in RAM can help reduce preparation time.  %
As for prefill and decode, %
the CPU can run at a higher frequency when a new inference process starts, ensuring the fastest response.

\vspace{0.05in} \noindent \revision{\textbf{Enhancing Memory Access Efficiency}
To address GPU underutilization, optimization efforts should primarily target memory access efficiency. Potential strategies include restructuring data layouts to improve locality, applying memory coalescing to reduce cache misses, and minimizing reliance on external memory. Furthermore, for long-context inference, partitioning large matrices into smaller chunks can effectively alleviate memory pressure and enhance execution efficiency.
}

\vspace{0.05in} \noindent \revision{\textbf{Customized Operators and Kernel Parameter Tuning}
Customized operator implementations are essential for fully exploiting GPU capabilities and achieving performance breakthroughs. Owing to the significant architectural differences across GPUs, generic operator implementations often fall short in delivering optimal acceleration. Fine-grained parameter tuning of kernel functions, such as adjusting the work size, offers a promising optimization path. For example, an excessively large global work size may introduce severe memory contention, while an overly small size can result in underutilization of compute units and additional overhead.
}

\vspace{0.05in} \noindent \revision{\textbf{Dynamic GPU Frequency}
The computing capability of the GPU is not fully utilized, as its operating frequency is generally fixed at a value lower than the maximum during inference. While this strategy helps maintain stable performance with moderate energy consumption, incorporating dynamic frequency adjustment could achieve a more adaptive balance between performance and energy efficiency.
}

\vspace{0.05in} \noindent \revision{\textbf{Hybrid Inference to Enhance User Experience}
 Although CPU performance appears sufficient, as shown in Section~\ref{subsec:thermal}, running inference solely on CPUs leads to rapid temperature increases. Given the strong performance of the NPU and GPU during the prefill stage, a hybrid inference approach—utilizing the NPU or GPU for prefill and the CPU for decoding—is expected to enhance efficiency while delaying temperature rise and improving the overall user experience.
}

\section{\revision{Concluding Remarks}}
\label{sec:conclusion}
In this paper, we present a comprehensive measurement study of LLM inference on mobile platforms, offering insights into both user-experience metrics (such as token throughput and energy consumption) and critical hardware and system characteristics (including CPU/GPU utilization, DVFS, and file mapping latency). Our analysis reveals how hardware capabilities and system dynamics impact on-device LLM performance and highlights potential bottlenecks affecting mobile LLM applications. Additionally, we propose potential directions for enhancing on-device LLM inference performance, \revision{including customizing operators by tuning kernel parameters to match various GPUs, optimizing data layout to improve memory efficiency, and searching for more adaptive DVFS policies to balance performance and energy.} We hope that this study can
provide insights for both the development of on-device LLMs and the design for future mobile system architecture.

\bibliographystyle{IEEEtran}
\bibliography{mine}{}

\vfill

\end{document}